\documentclass{article} 
\usepackage{iclr2027_conference,times}

\usepackage{amsmath,amsfonts,bm}

\def\eqref#1{equation~\ref{#1}}

\def\1{\bm{1}}

\DeclareMathAlphabet{\mathsfit}{\encodingdefault}{\sfdefault}{m}{sl}
\SetMathAlphabet{\mathsfit}{bold}{\encodingdefault}{\sfdefault}{bx}{n}

\usepackage{hyperref}
\usepackage{graphicx} 
\usepackage{url}
\usepackage{array}
\usepackage{makecell}
\usepackage{marvosym}

\usepackage{algorithm}
\usepackage{algorithmic}
\usepackage{multirow}
\usepackage{booktabs}
\usepackage{amsmath}
\usepackage{amssymb}
\usepackage[table]{xcolor}
\usepackage{colortbl} 
\usepackage[most]{tcolorbox}

\usepackage{wrapfig}
\usepackage{booktabs}

\usepackage{xcolor}
\usepackage{pifont}

\newcommand{\logo}{\raisebox{-4pt}{\includegraphics[width=1.8em]{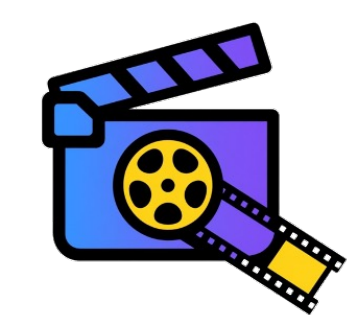}}}

\title{\logo~Beyond Coherence: Benchmarking Professional Editing-Technique Execution in Multi-Shot Audio-Video Generation}

\author{
Tianyi Zeng\textsuperscript{\rm 2}\thanks{Work done during internship at Alibaba}, 
Junchao Liao\textsuperscript{\rm 1}, 
Yujie Wei\textsuperscript{\rm 3}, 
Ziying Zhang\textsuperscript{\rm 1},
Litao Li\textsuperscript{\rm 1}, 
Tianyi Wang\textsuperscript{\rm 4}, \\
\textbf{
Zhichao Wei\textsuperscript{\rm 1},
Shuyao Xu\textsuperscript{\rm 1}, 
Wenwen Qiang\textsuperscript{\rm 5},
Siyu Zhu\textsuperscript{\rm 3},
Zhenghao Zhang\textsuperscript{\rm 1}\thanks{Project Leader \\ $\textsuperscript{\quad\ \ \Letter}$Corresponding Author}$\  \ ^\text{\Letter}$, 
Long Qin\textsuperscript{\rm 1}} \\
\textsuperscript{\rm 1}Alibaba Group \quad
\textsuperscript{\rm 2}Shanghai Jiao Tong University \quad
\textsuperscript{\rm 3}Fudan University\quad
\textsuperscript{\rm 4}UT Austin \\
\textsuperscript{\rm 5}Institute of Software, Chinese Academy of Sciences
\\
\texttt{zengtianyi@sjtu.edu.cn, zhangzhenghao.zzh@alibaba-inc.com}
}
\iclrfinalcopy 
\begin{document}

\maketitle

\begin{abstract}
Recent multi-shot audio-video generators can produce increasingly coherent and cinematic outputs, but coherence does not imply the ability to execute editing techniques. Professional editing depends on shot structure, transition grammar, audio-video cut relations, and montage, yet existing benchmarks largely rely on proxies such as content quality, synchronization, or physical plausibility, systematically missing whether such editing instructions are actually executed. We introduce \textbf{CutCraft}, the first benchmark for editing-technique execution in multi-shot audio-video generation. CutCraft extends structured multi-shot prompts with explicit editing specifications and is paired with a hierarchical hybrid evaluation framework that combines shot-structure alignment, expert-model metrics, tool-grounded multimodal judgment, and rubric-based question answering. Beyond evaluation, we design an agentic editing baseline that decomposes generation into planning, shot-level synthesis, and post-hoc composition, explicitly realizing editing semantics such as J-cuts, L-cuts, and transition timing. Across 13 state-of-the-art closed- and open-source models, CutCraft reveals a consistent gap between coherence and editing-technique execution: current systems often produce plausible multi-shot videos yet fail to execute editorial instructions reliably. We find unstable shot structures, weak control of transition execution, and sharp degradation on higher-order montage, while aesthetic quality is only weakly correlated with editing-technique compliance. The benchmark and metrics, and the editing agent baseline are available at \url{https://github.com/AlibabaResearch/cut-craft-bench}. 
\end{abstract}

\section{Introduction}

\begin{center}
\begin{minipage}{0.95\linewidth}
\centering
\itshape
``Two film pieces of any kind, placed together, inevitably combine into a new concept.''
\par\vspace{0.4em}
\raggedleft\normalfont
--- Sergei Eisenstein, \textit{The Film Sense}, 1942
\end{minipage}
\end{center}

\vspace{0.3em}

Recent advances in audio-video generation have made synthesized videos increasingly realistic, coherent, and cinematic \citep{seedance2026seedance,deepmind2025veo3,openai2025sora2,kuaishou2024Kling,wan2026wan27,liu2026javisdit++,hacohen2026ltx,low2025ovi,team2026mova}. Multi-shot generation, in particular, is beginning to resemble storytelling rather than isolated clip synthesis. But in video creation, coherence is only the surface. What gives a sequence structure, emphasis, and meaning is editing: how many shots appear, where cuts happen, how sound leads or lags the image, how adjacent shots continue, collide, or imply, and how a sequence is shaped into narrative, rhythm, or montage \citep{eisenstein1942film,martin1985langage,murch2001blink,bordwell2008film}. Editing techniques live in these decisions.

\begin{figure}[th!]
    \centering
    \includegraphics[width=1\linewidth]{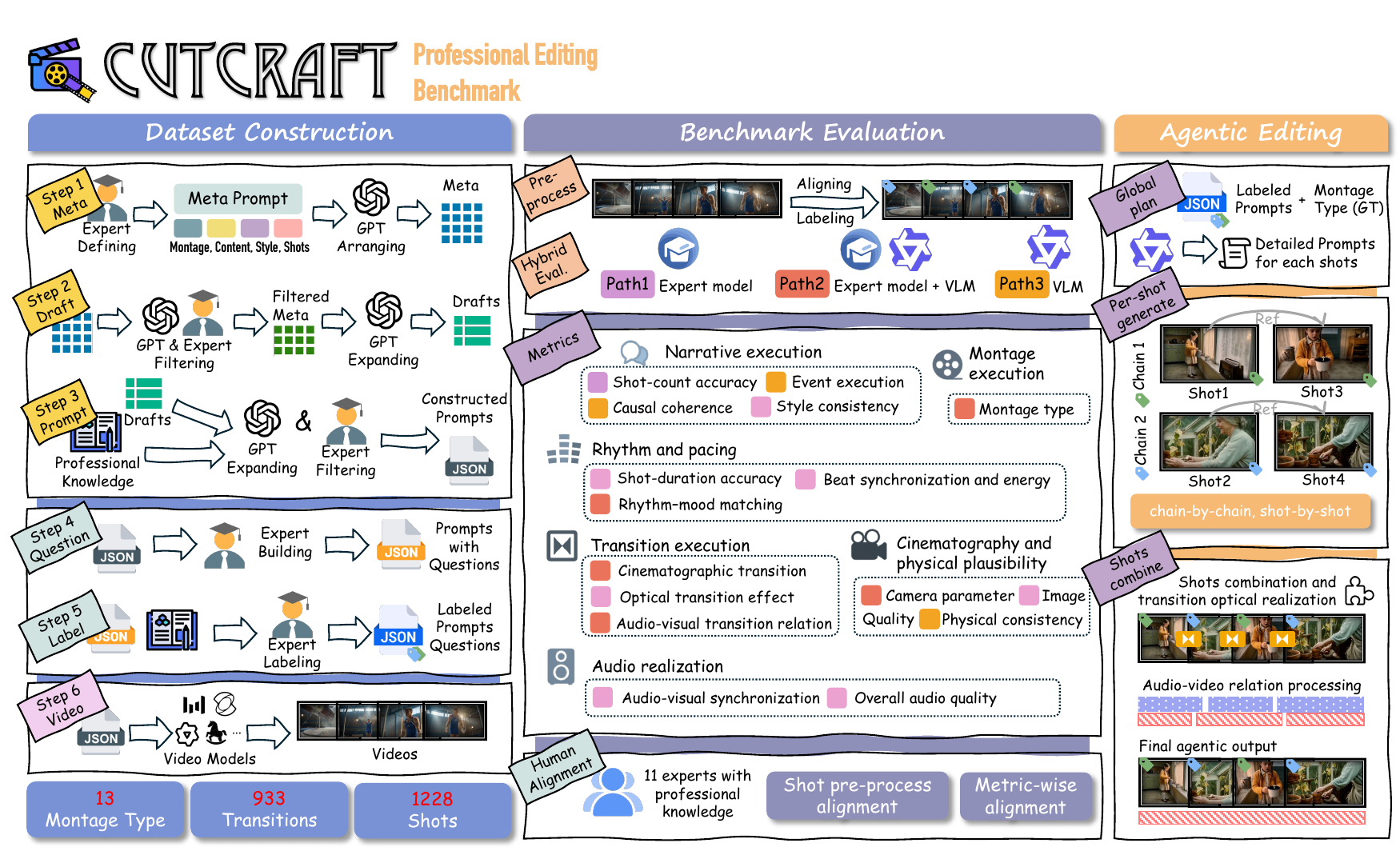}
    \caption{\textbf{CutCraft benchmark overview.} CutCraft targets professional editing-technique execution. It includes \textbf{(left)} an expert-curated data construction pipeline, \textbf{(middle)} a hierarchical editing-aware evaluation framework spanning narrative, rhythm, montage, transition, cinematography, and audio, and \textbf{(right)} an agentic editing baseline for controllable multi-shot generation.}
    \label{fig:placeholder}
\end{figure}

Current evaluation largely stops before this layer. A model may generate globally plausible multi-shot audio-video while still failing the very decisions that make a sequence editable. In practice, these failures recur in three forms: difficulty in realizing editing logic, especially for non-sequential montage involving parallel event lines or narrative discontinuities; unstable shot structuring, where the intended number and boundaries of shots cannot be reliably reproduced; and poor execution of fine-grained editing techniques, where transition effects degenerate into hard cuts and controlled audio-video relations such as J-cuts and L-cuts are rarely achieved. These are not cosmetic errors. They indicate that executing editing techniques is a distinct capability, not a byproduct of better rendering, stronger semantics, or smoother cross-shot coherence.

Existing benchmarks only partially cover this problem. Multi-shot audio-video benchmarks have significantly improved the evaluation of long-form generation, but largely treat editing structure as latent background rather than an object of measurement \citep{wei2026msavbench,liu2026longav,zhang2026multiref}. Fine-grained audio-video benchmarks probe semantic control, speech, or physical reasoning, but remain centered on single-shot or weakly structured settings \citep{zhou2026avgen,cui2026joint,xie2025phyavbench}. What remains missing is a benchmark for editing-technique execution in multi-shot audio-video generation.

To fill this gap, we introduce \textbf{CutCraft}, a benchmark for evaluating editing-technique execution in multi-shot audio-video generation. CutCraft extends structured multi-shot prompts with explicit editing specifications, including shot structure, transition type, audio-video cut relation, transition timing, and montage form.

This problem also requires a different evaluation design. Editing-aware assessment must handle structural mismatch between intended and generated shot layouts, distinguish synchronization from controlled asynchrony, and measure compliance across heterogeneous instruction types. We therefore propose a hierarchical hybrid evaluation framework built around three components: (i) a shot-alignment layer that resolves missing, merged, and fragmented shots before downstream scoring; (ii) a hybrid metric stack that combines expert-model measurements, tool-grounded multimodal judgment, and rubric-based question answering; and (iii) a task-specific compact MLLM judge, trained with On-Policy Distillation (OPD) \citep{li2026rethinking} for editing-aware evaluation, whose quality approaches strong closed-source evaluators while remaining lightweight.

Beyond evaluation, we also design an agentic editing baseline that decomposes multi-shot generation into planning, shot-level generation, and post-hoc composition. The agent parses the structured prompt into shot-level tasks, maintains cross-shot identity, setting, and sound consistency, generates short clips independently, and then realizes professional editing semantics during composition, including optical transitions, J-cuts, L-cuts, and audio timeline remixing. 

\begin{wrapfigure}{r}{0.4\textwidth}
    \centering
    \includegraphics[width=\linewidth]{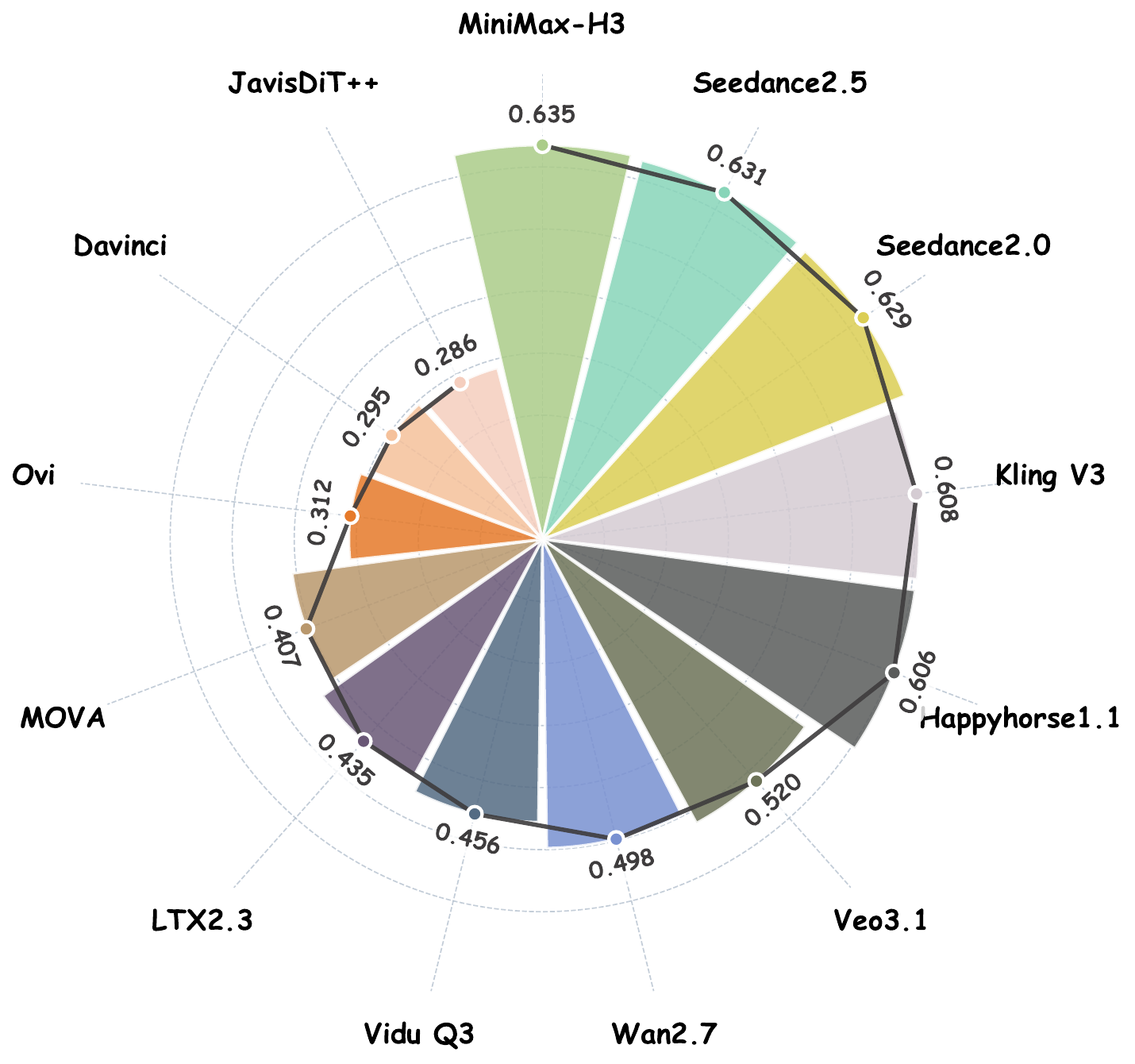}
    \caption{Overall statistical scores of video generation models on CutCraft.}
    \label{fig:radar}
\end{wrapfigure}

Once editing-technique execution is evaluated directly, a different picture emerges. Across 13 state-of-the-art closed- and open-source models \citep{seedance202625,minimax2026h3,wan2026wan27,happyhorse2026,kuaishou2024Kling,seedance2026seedance,deepmind2025veo3,bao2024vidu,hacohen2026ltx,low2025ovi,team2026mova,liu2026javisdit++,chern2026speed}, we find that current systems remain far closer to coherence than to editing-technique execution. Models that generate visually appealing and globally plausible content still fail to reliably follow editorial instructions. We further find a division of labor across architectures: modular or agentic systems tend to perform better on discrete editing controls such as shot duration or optical transition type, but worse on transitions that demand cross-shot spatial continuity, such as match cuts or occlusion-based wipes. Finally, aesthetic quality and editing-technique compliance are only weakly correlated, suggesting that existing benchmark scores may systematically obscure a central missing capability.
Our contributions are fourfold:
\begin{itemize}
\item We introduce CutCraft, the first benchmark dedicated to editing-technique execution in multi-shot audio-video generation, with structured prompts covering shot structure, transition relations, audio-video cut logic, and montage design.
\item We propose a hierarchical hybrid evaluation framework for editing-technique execution, centered on shot alignment, hybrid metrics, and a compact RL-trained MLLM judge.
\item We develop an agentic editing baseline that decomposes generation into planning, shot-level synthesis, and composition, and explicitly realizes professional editing semantics.
\item We provide a systematic empirical study of 13 representative state-of-the-art closed- and open-source systems, revealing fundamental limitations in shot-structure control, controlled audio-video asynchrony, montage execution, and the disconnect between aesthetics and editing-technique compliance.
\end{itemize}


We will release the benchmark and we hope CutCraft helps move the field from generating videos that merely look coherent to generating videos that are edited with intent.

\section{Related Work}
\paragraph{Benchmarks for video and audio-video generation.}
Benchmarking for video generation has expanded rapidly, with general-purpose benchmarks such as VBench~\citep{huang2024vbench}, VBench-2.0~\citep{zheng2025vbench}, EvalCrafter~\citep{liu2024evalcrafter}, FETV~\citep{liu2023fetv}, T2V-CompBench~\citep{sun2025t2v}, and Video-Bench~\citep{han2025video} evaluating perceptual quality, prompt alignment or compositionality. For audio-video generation, TAVGBench~\citep{mao2024tavgbench}, VABench~\citep{hua2026vabench}, and MSAVBench~\citep{wei2026msavbench} extend evaluation to synchronized or multi-shot audio-video outputs. Other recent benchmarks further probe finer-grained controllability or realism \citep{zhou2026avgen,cui2026joint,xie2025phyavbench,liu2026longav,zhang2026multiref}. These efforts substantially improve evaluation of generated content, but they mainly assess quality and coherence rather than whether professional editing instructions are actually executed.

\paragraph{Editing-oriented benchmarks.}
Several recent benchmarks move closer to cinematic editing. CineTechBench~\citep{wang2026cinetechbench} studies cinematographic techniques, while VEBench~\citep{deng2026vebench}, VEU-Bench~\citep{li2025veu}, and related efforts examine video editing understanding, editing workflows, or creation-and-editing reasoning. ViStoryBench~\citep{zhuang2026vistorybench} further evaluates story-level consistency and narrative structure. These works are valuable for measuring cinematic literacy and editing understanding, but they primarily focus on recognizing, analyzing, or reasoning about editing in existing videos. In contrast, our focus is on whether generative models can execut editing technique in synthesized multi-shot audio-video outputs. To our knowledge, CutCraft is the first benchmark designed explicitly for this setting.

\section{CutCraft}

\subsection{Data Design}

\subsubsection{Data Construction}

The \textbf{CutCraft dataset} is built through a multi-stage pipeline that combines LLM-assisted drafting with repeated expert review. The taxonomy and terminology are expert-defined, and each major stage is manually reviewed to keep the final samples suitable for evaluating editing-technique execution.

We begin from a meta space defined over four attributes: montage subtype, video content category, visual style, and number of shots. An LLM expands this space into candidate meta combinations, which are then screened and deduplicated to remove repetitive, implausible, or weak cases. Each curated meta prompt is expanded into a 15-second multi-shot draft with a global description and shot-level descriptions. Drafting is constrained by shot count, temporal coverage, single-shot continuity, and plausible audio-video events.
The drafts are then converted into structured prompts using an expert-defined vocabulary of montage attributes and transition types. Each shot is associated with camera-related specifications, and each shot boundary with a transition specification. 

From the finalized prompts, human experts build a question bank for evaluation. Each sample is paired with six core questions covering event order, causal coherence, montage recognition, physical consistency, camera-parameter recognition, and transition recognition. 

Finally, we attach montage-related labels at shot levels. Each sample inherits a fine-grained montage subtype from the taxonomy, and each shot is assigned an \texttt{event\_coherence\_label} indicating its event chain grouping under the intended montage structure. Together, these prompts, questions, and labels form the basis of the CutCraft benchmark, the details of data construction are provided in Appendix \ref{app:data_construction}.

\subsubsection{Data Analysis}

\begin{figure}[t!]
    \centering
    \includegraphics[width=1\linewidth]{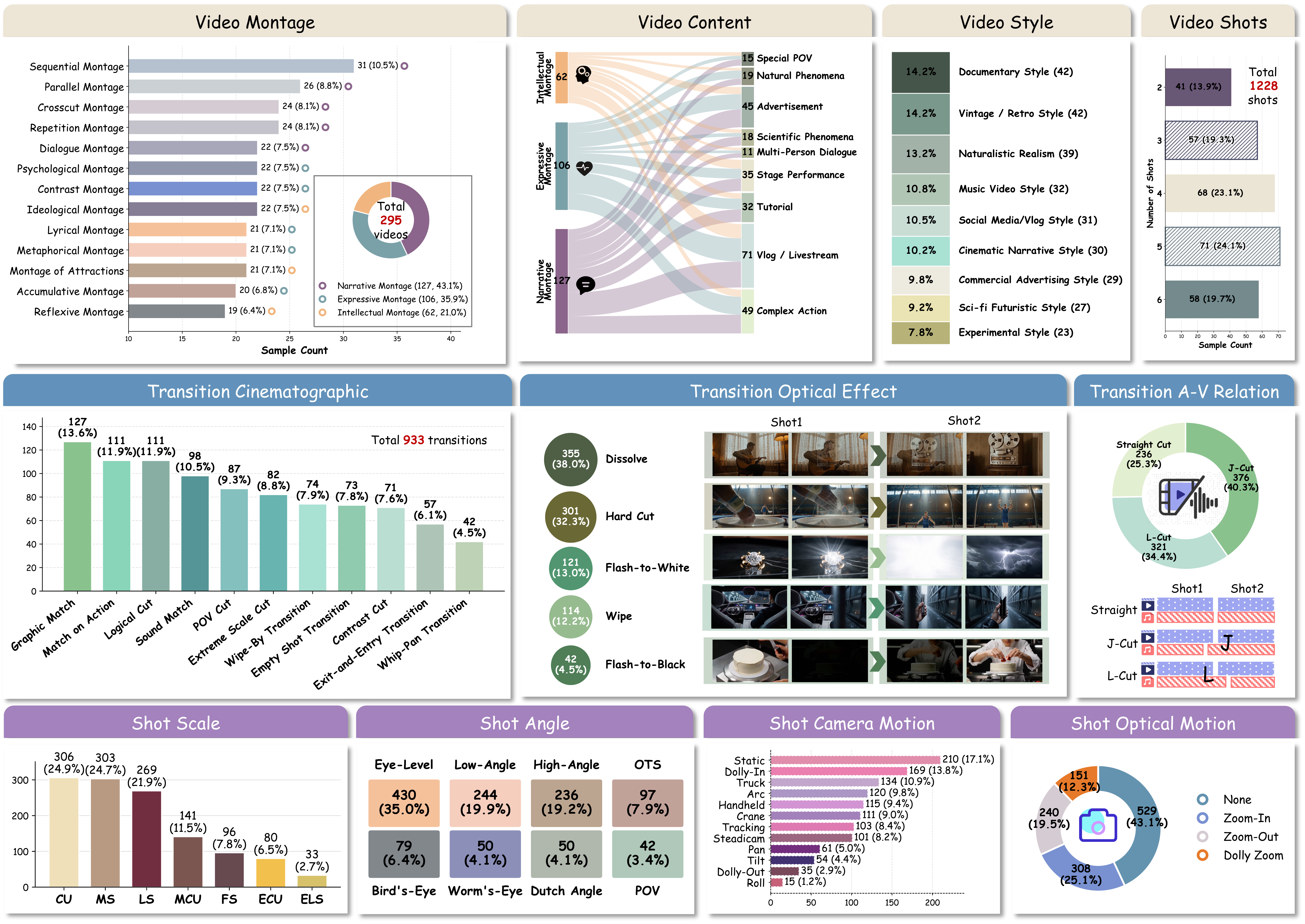}
    \caption{\textbf{Data distribution of CutCraft.} CutCraft maintains broad coverage across the main dimensions of professional editing, including video-level, transition-level and shot-level attributes.}
    \label{fig:data_ditribution}
\end{figure}

The \textbf{CutCraft dataset} is produced through strict multi-stage filtering. Starting from \textbf{1500} LLM-generated candidate meta prompts, expert screening and deduplication retain \textbf{534}. These are then expanded into \textbf{534} textual drafts. After further filtering, normalization, and expert revision, the final release contains \textbf{295} structured prompt samples, corresponding to \textbf{19.7\%} of the initial candidate pool. We further construct \textbf{1770} core evaluation questions and \textbf{933} transition-related survey questions. This high attrition rate reflects a design choice: prioritizing curation quality over raw scale.

The final dataset contains \textbf{295} samples, \textbf{1228} shots, and \textbf{933} inter-shot transitions, and it combines strict expert curation with broad professional diversity. \textbf{Montage diversity.} The dataset covers all \textbf{13} target montage subtypes, spanning narrative, expressive, and intellectual montage. \textbf{Content diversity.} Samples span a wide range of content domains, visual styles, and shot counts, covering both everyday scenarios and professionally challenging compositions. \textbf{Transition diversity.} The dataset exhibits broad variation in cinematographic transition types, optical effects, and audio-video transition relations. \textbf{Shot-level cinematography diversity.} At the shot level, the dataset spans diverse cinematic parameters, including shot scale, camera angle, camera motion, optical motion, and depth-related settings.
Detailed category-wise statistics are provided in Appendix \ref{app:data_analysis}.

\subsection{Evaluation}

A generated video may appear coherent, visually appealing, and broadly aligned with the prompt, yet still fail the editorial decisions that define the sequence: how many shots appear, where cuts happen, whether sound leads or lags the image, and whether adjacent shots realize the intended montage or transition logic. This creates three methodological challenges. First, generated videos often deviate from the prompt-defined shot layout, so many editing metrics are ill-posed without structural alignment. Second, different editing dimensions require different kinds of evidence: some depend on direct temporal or signal measurements, while others require multimodal reasoning over cross-shot structure. Third, several core editing constructs, such as montage and J/L-cuts, cannot be reduced to conventional proxies like coherence or synchronization.
We address these challenges with the following metrics and the hierarchical evaluation framework design. 

\subsubsection{Evaluation Metrics Design}

Our metrics are organized into six groups, which together cover the main components of professional editing.
\textbf{A. Narrative execution.}
The A group measures whether the generated video realizes the intended multi-shot narrative structure. \textbf{A1} evaluates shot-count accuracy. \textbf{A2} measures shot-by-shot event execution alignment. \textbf{A3} evaluates causal coherence between adjacent or related shots. \textbf{A4} measures style consistency.
\textbf{B. Rhythm and pacing.}
The B group evaluates temporal organization at the editing level. \textbf{B1} measures shot-duration accuracy relative to the intended pacing. \textbf{B2} evaluates beat synchronization through cut timing, motion-energy variation, and audio energy. \textbf{B3} measures rhythm-mood matching, i.e., whether the editing pace and soundtrack emotion fit the intended scene. 
\textbf{C. Montage execution.} \textbf{C1} measures whether the generated video instantiates the intended montage subtype, one of the central targets of CutCraft.
\textbf{D. Transition execution.}
The D group evaluates inter-shot transitions at three levels. \textbf{D1} measures cinematographic transition design. \textbf{D2} evaluates optical transition effects. \textbf{D3} evaluates audio-video transition relations. 
\textbf{E. Cinematography and physical plausibility.}
The E group focuses on shot-level visual execution. \textbf{E1} evaluates camera parameters. \textbf{E2} measures image quality. \textbf{E3} evaluates physical consistency. 
\textbf{F. Audio realization.}
The F group evaluates audio execution. \textbf{F1} measures within-shot audio-video synchronization. \textbf{F2} evaluates overall audio quality in terms of signal quality, content richness, and perceptual naturalness. 
More details are provided in Appendix \ref{app:metrics_design}.

\subsubsection{Hierarchical Evaluation Framework}


\paragraph{Shot alignment and structural labeling.}
In practice, generated videos often fail to preserve the prompt-defined shot layout: intended shots may be missing, fragmented into multiple segments, or merged together. Without resolving this mismatch first, many editing metrics become ambiguous or invalid.
So, we first apply an automatic shot-boundary detector~\citep{soucek2024transnet} to segment the generated video into raw temporal units. A vision-language model~\citep{team2026qwen3} then aligns these segments to the ground-truth shot plan. The alignment procedure determines whether each prompt shot is \texttt{matched}, \texttt{merged}, or \texttt{missing}, and includes post-processing to correct non-adjacent merges, label jumps, and temporal inconsistencies.

\paragraph{Three evidence pathways.}
On top of this structure, we build an evidence-grounded metric suite with three complementary scoring pathways.
More details are provided in Appendix \ref{app:metrics_design}.

\textbf{Path-I: direct expert-model scoring.}
This pathway is used when the target can be measured from explicit temporal or signal evidence~\citep{soucek2024transnet,teed2020raft,zhang2025monst3r,wang2024dust3r,radford2021learning,cherti2023reproducible,oquab2023dinov2,schuhmann2022laion,huang2024vbench,rao2020unified,redmon2016you,viola2001rapid,hou2007saliency,pech2000diatom,hempel20226d,rouard2023hybrid,radford2023robust,kong2020panns}. Representative examples include \textbf{B1} ({shot-duration accuracy}), which compares aligned shot durations against the prompt-defined pacing plan, and \textbf{D2} ({optical transition type}), which classifies whether a shot boundary realizes the annotated visual effect, such as hard cut, dissolve, wipe, flash-to-white, or flash-to-black. These metrics are grounded in measurable evidence and therefore provide reliable low-level estimates of editing execution.

\begin{figure}[t!]
    \centering
    \includegraphics[width=1\linewidth]{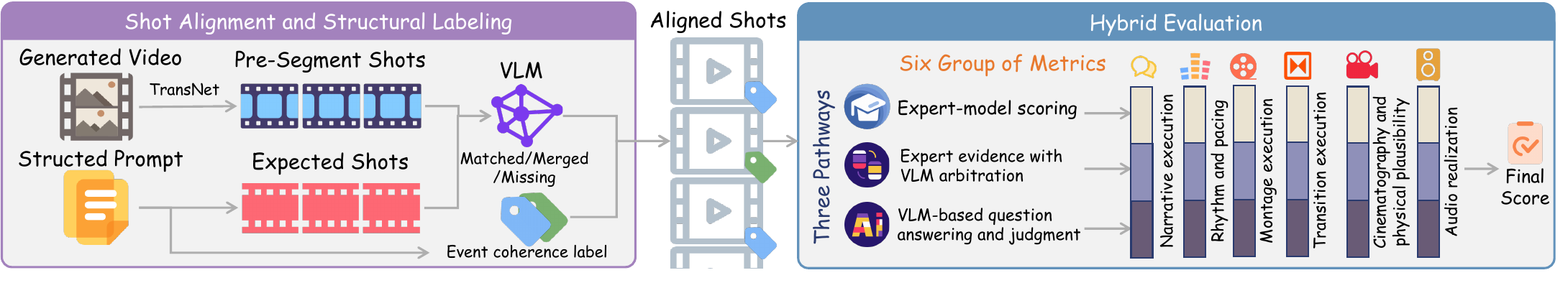}
    \caption{\textbf{Hierarchical evaluation framework of CutCraft.} Generated videos are first aligned to the prompt-defined shot structure and enriched with structural labels. A hybrid evaluation suite then scores six editing-aware metric groups using three evidence pathways: direct expert-model scoring, expert evidence with VLM arbitration, and VLM-based judgment.}
    \label{fig:eval_frame}
\end{figure}

\textbf{Path-II: expert evidence with VLM arbitration.}
This pathway is used when low-level evidence is informative but insufficient without semantic interpretation. A central example is \textbf{C1} ({montage execution}), where we combine shot-grouping signals, cross-shot similarity evidence, and prompt-side event-coherence labels to support a VLM decision over the target montage subtype. 
For example, \textbf{D3} ({audio-video transition relation}) evaluates whether a boundary realizes the annotated J-cut, L-cut, or straight cut by combining temporal audio evidence around the cut with VLM judgment over the corresponding transition clip.
This pathway captures whether the generated video realizes the intended editing language of shot connection.

\textbf{Path-III: VLM-based question answering and judgment.}
This pathway is used for dimensions that are inherently semantic, reasoning-heavy, or defined at the narrative-editing level. For example, \textbf{A2} ({event execution alignment}) evaluates whether the video follows the prompt as a shot-wise execution plan: the evaluator first identifies the realized event chain, then checks shot presence through alignment, and finally judges whether each intended event is fully, partially, or poorly realized. 
This pathway is especially important for constructs that cannot be reduced to direct low-level measurements, including controlled audio-video asynchrony and higher-level editing logic.


\subsection{Agentic Editing Baseline}

To test whether editing-technique execution benefits from editing-aware inference, the agentic baseline uses a plan-generate-compose-repair pipeline instead of one-shot 15-second generation. It decomposes each structured prompt into shot-level tasks, performs global consistency analysis over labels such as \texttt{event\_coherence\_label}, and injects shared subject, setting, and event-line constraints into each shot prompt. Shots with the same label reuse earlier visual references, and later shots follow logical event continuity. Each shot is generated with extra headroom so that transition overlap and audio offsets can be executed during composition.
Composition explicitly maps editing annotations to rendering operations. Optical transitions are implemented as cuts, dissolves, wipes, or flash effects, while J-cuts and L-cuts are realized on a separate audio timeline by offsetting the audio. After composition, the result is evaluated on the same editing dimensions as CutCraft. A central planner attributes failures to specific transitions and applies targeted fixes, such as adjusting transition duration, cut timing, or audio offsets and shot prompts, by either re-composing the video or re-generating only affected shots for up to two rounds.
More details of the agentic baseline are provided in Appendix \ref{app:agentic_baseline}.

\section{Experiments}

\subsection{Experimental Setup}

We evaluated eight closed-source commercial models, including Seedance  2.5\citep{seedance202625}, Minimax H3 \citep{minimax2026h3}, Seedance 2.0 \citep{seedance2026seedance}, Happyhorse 1.1 \citep{happyhorse2026}, Kling V3 \citep{kuaishou2024Kling}, Wan 2.7 \citep{wan2026wan27}, Veo 3.1 \citep{deepmind2025veo3} and Vidu Q3 \citep{bao2024vidu}, as well as five open-source models, including LTX 2.3 \citep{hacohen2026ltx}, MOVA \citep{team2026mova}, Ovi \citep{low2025ovi}, Davinci \citep{chern2026speed} and JavisDiT++ \citep{liu2026javisdit++}. In addition, we applied our agentic framework to Happyhorse 1.1 \citep{happyhorse2026}, Wan 2.7 \citep{wan2026wan27} and LTX 2.3 \citep{hacohen2026ltx}. 
More details of experimental settings are provided in Appendix \ref{app:implementation}.

\subsection{Results and Analysis}

\begin{table*}[ht!]
\centering
\caption{Main results on CutCraft across closed-source, open-source, and agentic generation settings. Current models remain substantially stronger on coherence-related dimensions than on editing-specific dimensions such as montage and transition execution.}
\label{tab:main_result}
\resizebox{\textwidth}{!}{%
\begin{tabular}{l|cccc|ccc|c|ccc|ccc|cc|c}
\toprule

\multirow{2}*{\textbf{Model}} &  \multicolumn{4}{c|}{\textbf{Narrative}} & \multicolumn{3}{c|}{\textbf{Rhythm \& pacing}} & \multicolumn{1}{c|}{\textbf{Montage}} & \multicolumn{3}{c|}{\textbf{Transition}} &
\multicolumn{3}{c|}{\textbf{Cinematography}} & \multicolumn{2}{c|}{\textbf{Audio}} & \multirow{2}*{\textbf{Overall $\uparrow$}} \\

~ & A1$\uparrow$ & A2$\uparrow$ & A3$\uparrow$ & A4$\uparrow$ & B1$\uparrow$ & B2$\uparrow$ & B3$\uparrow$ & C1$\uparrow$ & D1$\uparrow$ & D2$\uparrow$ & D3$\uparrow$ & E1$\uparrow$ & E2$\uparrow$ & E3$\uparrow$ & F1$\uparrow$ & F2$\uparrow$ & ~ \\
\midrule

\rowcolor{gray!20}\multicolumn{18}{l}{\textit{Closed-source models}} \\

Minimax H3 & \cellcolor{cyan!20}0.948 & \cellcolor{cyan!40}0.815 & \cellcolor{cyan!40}0.729 & 0.636 & \cellcolor{cyan!40}0.938 & \cellcolor{cyan!10}0.563 & \cellcolor{cyan!10}0.677 & \cellcolor{cyan!40}0.502 & \cellcolor{cyan!10}0.406 & 0.463 & \cellcolor{cyan!40}0.394 & \cellcolor{cyan!40}0.592 & \cellcolor{cyan!20}0.558 & \cellcolor{cyan!20}0.847 & 0.603 & 0.526 & \cellcolor{cyan!40}0.635 \\

Seedance2.5 & \cellcolor{cyan!10}0.884 & \cellcolor{cyan!10}0.766 & 0.705 & \cellcolor{cyan!20}0.660 & \cellcolor{cyan!10}0.848 & \cellcolor{cyan!40}0.581 & \cellcolor{cyan!20}0.686 & \cellcolor{cyan!20}0.498 & \cellcolor{cyan!20}0.421 & \cellcolor{cyan!10}0.511 & \cellcolor{cyan!10}0.338 & 0.544 & \cellcolor{cyan!10}0.544 & \cellcolor{cyan!10}0.845 & 0.745 & 0.538 & \cellcolor{cyan!20}0.631 \\

Seedance2.0 & 0.865 & \cellcolor{cyan!20}0.800 & \cellcolor{cyan!10}0.712 & \cellcolor{cyan!10}0.653 & 0.739 & \cellcolor{cyan!20}0.571 & 0.653 & \cellcolor{cyan!10}0.490 & \cellcolor{cyan!40}0.440 & \cellcolor{cyan!40}0.584 & \cellcolor{cyan!20}0.382 & \cellcolor{cyan!20}0.561 & 0.539 & 0.829 & 0.687 & \cellcolor{cyan!20}0.555 & \cellcolor{cyan!10}0.629 \\

Kling V3 & \cellcolor{cyan!40}0.956 & 0.737 & \cellcolor{cyan!20}0.716 & \cellcolor{cyan!10}0.653 & \cellcolor{cyan!20}0.911 & 0.523 & 0.633 & 0.454 & 0.405 & 0.476 & 0.305 & \cellcolor{cyan!10}0.554 & 0.527 & 0.834 & 0.480 & \cellcolor{cyan!40}0.560 & 0.608 \\

Happyhorse1.1 & 0.853 & 0.643 & 0.689 & 0.635 & 0.822 & 0.541 & \cellcolor{cyan!40}0.688 & 0.458 & 0.383 & \cellcolor{cyan!20}0.554 & 0.311 & 0.516 & 0.536 & \cellcolor{cyan!40}0.868 & 0.673 & 0.533 & 0.606 \\

Veo3.1 & 0.757 & 0.441 & 0.512 & 0.638 & 0.581 & 0.504 & 0.608 & 0.371 & 0.267 & 0.320 & 0.237 & 0.393 & \cellcolor{cyan!40}0.560 & 0.799 & \cellcolor{cyan!10}0.792 & \cellcolor{cyan!10}0.546 & 0.520 \\

Wan2.7 & 0.802 & 0.451 & 0.549 & \cellcolor{cyan!40}0.666 & 0.556 & 0.452 & 0.543 & 0.431 & 0.305 & 0.355 & 0.231 & 0.432 & 0.504 & 0.737 & 0.436 & 0.512 & 0.498 \\

Vidu Q3 & 0.637 & 0.398 & 0.432 & 0.604 & 0.306 & 0.529 & 0.557 & 0.391 & 0.214 & 0.257 & 0.190 & 0.356 & 0.515 & 0.633 & 0.738 & 0.537 & 0.456 \\

\midrule
\rowcolor{gray!20}\multicolumn{18}{l}{\textit{Open-source models}} \\

LTX2.3 & 0.631 & 0.291 & 0.391 & 0.642 & 0.244 & 0.545 & 0.530 & 0.334 & 0.200 & 0.190 & 0.186 & 0.336 & 0.529 & 0.727 & 0.667 & 0.516 & 0.435 \\

MOVA & 0.578 & 0.151 & 0.296 & 0.545 & 0.479 & 0.454 & 0.487 & 0.263 & 0.154 & 0.206 & 0.157 & 0.272 & 0.461 & 0.658 & \cellcolor{cyan!20}0.858 & 0.494 & 0.407 \\

Ovi & 0.417 & 0.078 & 0.081 & 0.560 & 0.160 & 0.367 & 0.386 & 0.222 & 0.034 & 0.036 & 0.030 & 0.145 & 0.541 & 0.556 & \cellcolor{cyan!40}0.860 & 0.517 & 0.312 \\

Davinci & 0.321 & 0.094 & 0.029 & 0.579 & 0.019 & 0.354 & 0.584 & 0.188 & 0.004 & 0.010 & 0.006 & 0.158 & 0.534 & 0.578 & 0.730 & 0.534 & 0.295 \\

JavisDiT++ & 0.431 & 0.059 & 0.062 & 0.494 & 0.470 & 0.344 & 0.451 & 0.229 & 0.037 & 0.044 & 0.041 & 0.147 & 0.428 & 0.437 & 0.529 & 0.364 & 0.286 \\

\midrule
\rowcolor{gray!20}\multicolumn{18}{l}{\textit{Models with agentic generation}} \\

Happyhorse1.1* & 0.890 & 0.700 & 0.684 & 0.644 & 0.864 & \cellcolor{yellow!30}0.556 & \cellcolor{yellow!30}0.635 & \cellcolor{yellow!30}0.493 & 0.381 & 0.659 & 0.555 & \cellcolor{yellow!30}0.553 & \cellcolor{yellow!30}0.536 & 0.870 & \cellcolor{yellow!30}0.479 & \cellcolor{yellow!30}0.588 & \cellcolor{yellow!30}0.630 \\
Wan2.7* & \cellcolor{yellow!30}0.896 & \cellcolor{yellow!30}0.748 & \cellcolor{yellow!30}0.699 & \cellcolor{yellow!30}0.658 & \cellcolor{yellow!30}0.913 & 0.553 & 0.541 & 0.464 & \cellcolor{yellow!30}0.417 & \cellcolor{yellow!30}0.666 & \cellcolor{yellow!30}0.619 & 0.547 & 0.519 & \cellcolor{yellow!30}0.871 & 0.277 & 0.581 & 0.623 \\
LTX2.3* & 0.875 & 0.613 & 0.676 & \cellcolor{yellow!30}0.658 & 0.807 & 0.531 & 0.563 & 0.442 & 0.376 & 0.510 & 0.512 & 0.525 & 0.509 & 0.816 & 0.477 & 0.570 & 0.591 \\

\bottomrule
\end{tabular}%
}
\end{table*}

Table~\ref{tab:main_result} and Figure \ref{fig:radar} report the main results on CutCraft. Across all models, a consistent gap emerges between coherence and editing-technique execution.

\begin{figure}[ht!]
    \centering
    \includegraphics[width=1\linewidth]{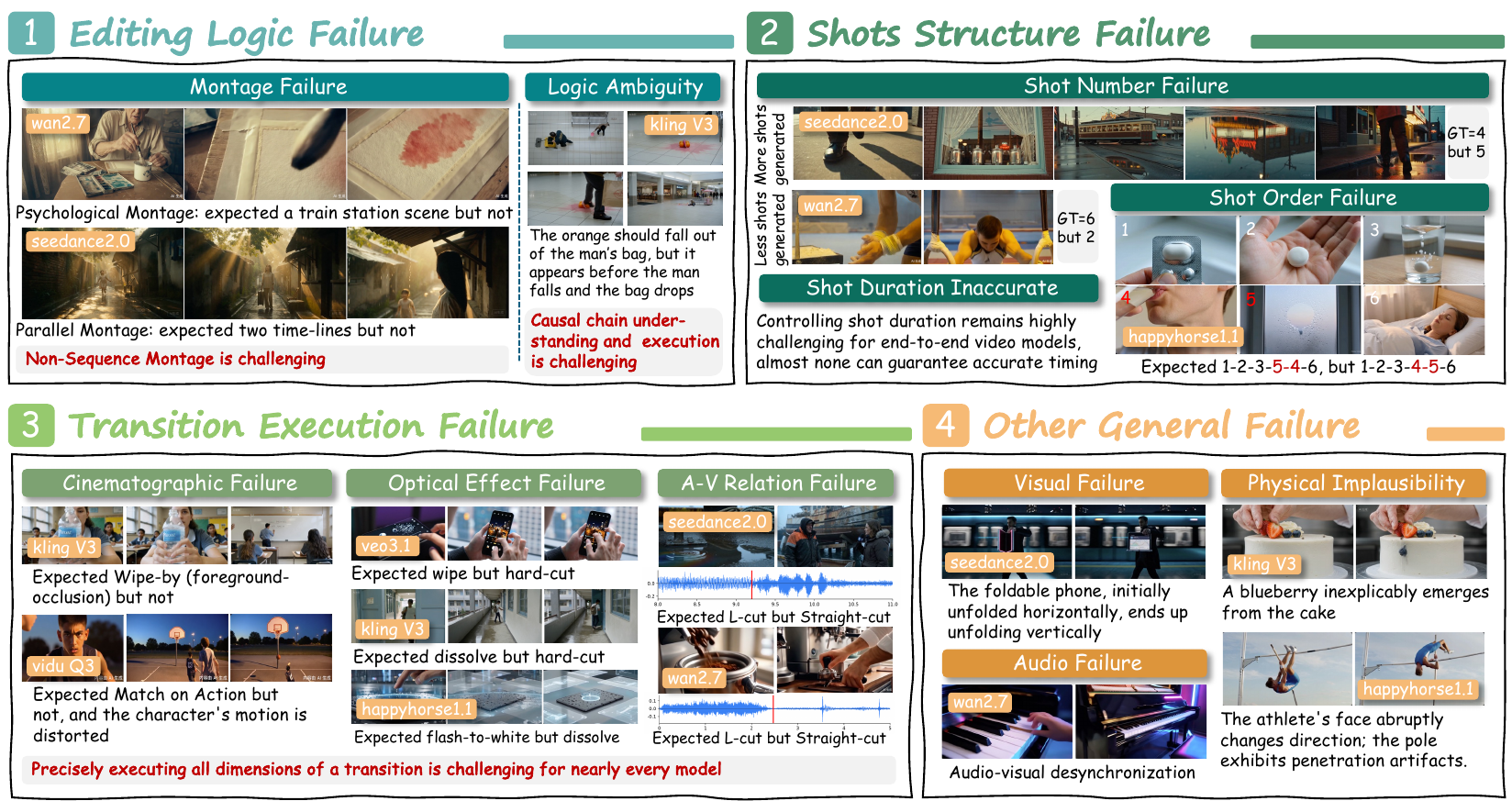}
    \caption{\textbf{Representative failure cases on CutCraft.} The examples show four common failure modes of current models: editing-logic, shot-structure, transition-execution, and general failures. }
    \label{fig:case_study}
\end{figure}

\textbf{(i) Discontinuous montage remains a major failure mode.} Table~\ref{tab:montage} provide fine-grained quantitative analyses, and Figure~\ref{fig:case_study} shows representative qualitative examples; additional cases are deferred to Appendix~\ref{app:more_cases}. As shown in Table~\ref{tab:montage}, all models achieve higher montage execution scores (\textbf{C1}) on montage types with continuous event chains than on those with discontinuous event chains. 
\begin{wraptable}{r}{0.43\textwidth}
    \vspace{-1em}
    \centering
    \caption{Visual quality (\textbf{E2}) is only weakly correlated with editing-related dimensions.}
    \resizebox{0.43\textwidth}{!}{%
    \begin{tabular}{l|ccccccc}
    \toprule
        \textbf{E2} vs. & \textbf{A1} & \textbf{A2} & \textbf{A3} & \textbf{C1} & \textbf{D1} & \textbf{D2} & \textbf{D3} \\
    \midrule

    Spearman & 0.34 & 0.49 & 0.42 & 0.40 & 0.40 & 0.32 & 0.49\\

    p value & 0.25 & 0.09 & 0.16 & 0.17 & 0.17 & 0.29 & 0.09\\

    \bottomrule
    \end{tabular}%
    }
    \label{tab:cor}
\end{wraptable}
\textbf{(ii) Transition execution degrades with shot complexity.}
Transition execution remains another major challenge. Table~\ref{tab:shots_D} reports overall transition performance, averaged over \textbf{D1}, \textbf{D2}, and \textbf{D3}, grouped by shot count. Performance generally degrades as the number of shots increases, indicating that maintaining transition control becomes harder as multi-shot structure grows more complex. 
\textbf{(iii) Agentic generation improves discrete editorial control.}
The agentic baseline partially closes this gap. As shown in the last three rows of Table~\ref{tab:main_result}, agentic generation substantially improves dimensions tied to explicit planning and post-hoc composition, particularly shot structure (\textbf{A1}, \textbf{A2}), pacing control (\textbf{B1}), and optical transition execution (\textbf{D2}). 
\textbf{(iv) Aesthetic quality is a weak proxy for editing-technique execution.}
Finally, visual quality is only weakly related to editing-technique execution. In practice, several models~\citep{hacohen2026ltx,low2025ovi,chern2026speed} score relatively well on image quality (\textbf{E2}) while still performing poorly on editing-related dimensions. Table~\ref{tab:cor} confirms this quantitatively: the correlation between \textbf{E2} and editing dimensions remains weak.

\begin{table*}[ht!]
    \centering

    \begin{minipage}[t]{0.52\textwidth}
    \centering
    \caption{Montage execution is substantially harder for montage types with discontinuous event chains.}
    \resizebox{\textwidth}{!}{
    \begin{tabular}{l|cc|cc}
    \toprule
    \multirow{2}*{\textbf{Method}}
    & \multicolumn{2}{c|}{\textbf{\makecell{Montage with\\Continuous Chain}}}
    & \multicolumn{2}{c}{\textbf{\makecell{Montage with\\Discontinuous Chain}}} \\
    ~ & C1 & Overall & C1 & Overall \\
    \midrule

    Minimax H3 & 0.584 & 0.640 & 0.428 & 0.631 \\
    Seedance2.5 & 0.575 & 0.623 & 0.432 & 0.638 \\
    Seedance2.0 & 0.576 & 0.629 & 0.414 & 0.628 \\
    Kling V3 & 0.500 & 0.614 & 0.414 & 0.602 \\
    Happyhorse1.1 & 0.533 & 0.611 & 0.392 & 0.602 \\
    Veo3.1 & 0.478 & 0.514 & 0.277 & 0.526 \\
    Wan2.7 & 0.507 & 0.500 & 0.363 & 0.496 \\
    Vidu Q3 & 0.467 & 0.453 & 0.324 & 0.458 \\
    LTX2.3 & 0.453 & 0.446 & 0.229 & 0.424 \\
    MOVA & 0.446 & 0.415 & 0.102 & 0.400 \\
    Ovi & 0.366 & 0.315 & 0.096 & 0.309 \\
    Davinci & 0.355 & 0.311 & 0.041 & 0.281 \\
    JavisDiT++ & 0.362 & 0.297 & 0.111 & 0.276 \\

    \bottomrule
    \end{tabular}
    }
    \label{tab:montage}
\end{minipage}
    \hfill
    \begin{minipage}[t]{0.44\textwidth}
        \centering
        \caption{Transition execution degrades as the number of shots increases. Scores are averaged over \textbf{D1}, \textbf{D2}, and \textbf{D3}.}
        \resizebox{\textwidth}{!}{
        \begin{tabular}{l|ccc}
        \toprule
            \textbf{Method} & \textbf{2 shots} & \textbf{3--4 shots} & \textbf{5--6 shots} \\
        \midrule
        
        Minimax H3 & 0.542 & 0.421 & 0.383 \\
        Seedance2.5 & 0.466 & 0.417 & 0.416 \\
        Seedance2.0 & 0.522 & 0.480 & 0.441 \\
        Happyhorse1.1 & 0.494 & 0.432 & 0.376 \\
        Kling V3 & 0.454 & 0.387 & 0.385 \\
        Veo3.1 & 0.424 & 0.292 & 0.211 \\
        Wan2.7 & 0.469 & 0.323 & 0.217 \\
        Vidu Q3 & 0.401 & 0.243 & 0.140 \\
        LTX2.3 & 0.405 & 0.208 & 0.108 \\
        MOVA & 0.156 & 0.189 & 0.162 \\
        Ovi & 0.084 & 0.036 & 0.014 \\
        Davinci & 0.033 & 0.004 & 0.001 \\
        JavisDiT++ & 0.079 & 0.034 & 0.035 \\
        \bottomrule
        \end{tabular}
        }
        \label{tab:shots_D}
    \end{minipage}

\end{table*}

\subsection{Key Findings}


\textbf{Finding 1: Editing-technique execution is not implied by coherence.}
Many current models can generate videos that are visually plausible, temporally smooth, and broadly aligned with the prompt, yet still fail to execute the intended editing plan.

\textbf{Finding 2: Current models are biased toward continuity-preserving generation.}
Models perform substantially better on editing patterns with continuous event structure than on montage types that require discontinuous event organization across shots. 
Together, these results suggest that current systems favor local continuity over higher-order editorial logic.

\textbf{Finding 3: Explicit decomposition helps discrete editorial control more than higher-order editing logic.}
The agentic baseline improves dimensions that can be operationalized through explicit planning and composition, such as shot count, pacing, and optical transition execution. 
However, the gains are smaller on dimensions that require more semantic or rhetorical organization. This suggests that editing-aware decomposition is helpful, but does not fully solve editing-technique execution.

\textbf{Finding 4: Visual quality is a poor proxy for editing-technique execution.}
Visual aesthetics and editing execution are only weakly correlated. A model can produce high-quality footage while still failing at shot organization, montage, or transition control. This result highlights the need for benchmarks that evaluate editing-specific capabilities directly.



\subsection{Human Validation of Metrics and Evaluator Selection}

We validate the effectiveness of six key editing-aware metrics by comparing automated scores against expert judgments. 
To this end, we recruited eleven experts with backgrounds in film and video editing to manually assess \textbf{640} video samples, consisting of 80 randomly sampled videos from each of 8 closed-source models.
The details are provided in Appendix~\ref{app:human_aligment}.
We also compare two candidate MLLM evaluators, Qwen3.5-Omni and Qwen3-VL-Plus, by measuring their correlation with human ratings. Table~\ref{tab:human_align} reports the results. 
Across the dimensions, Qwen3.5-Omni shows stronger alignment with expert judgment, with an average Pearson correlation of 0.92 and an average Spearman correlation of 0.91. 
These results support the validity of our metric design and motivate the use of Qwen3.5-Omni as the evaluator.
We further train a compact evaluator for practical deployment. Specifically, we apply OPD-based post-training \citep{li2026rethinking} to Qwen3-VL-8B. As shown in Table~\ref{tab:human_align1}, the model achieves performance close to Qwen3.5-Omni on test set. 
Training details are deferred to Appendix \ref{app:OPD}.

\begin{table*}[ht!]
    \centering
    \begin{minipage}[t]{0.56\textwidth}
        \centering
        \caption{Correlation between human ratings and automated evaluation scores.}
        \resizebox{1\textwidth}{!}{%
        \begin{tabular}{c|c|cccccc}
        \toprule
        Correlation& VLM  & A2 & A3 & C1 & D1 & D3 & E3 \\
        \midrule
        \multirow{2}*{Pearson $\uparrow$}  & Qwen3.5-Omni  & 0.94 & 0.93 & 0.85 & 0.92 & 0.97 & 0.89 \\
         ~ & Qwen3-VL-Plus  & 0.87 & 0.92 & 0.77 & 0.70 & 0.81 & 0.61\\
    
        \midrule
        
        \multirow{2}*{Spearman $\uparrow$} & Qwen3.5-Omni  & 1.00 & 0.80 & 0.80 & 0.95 & 0.95 & 0.95 \\
        ~ & Qwen3-VL-Plus & 1.00 & 0.32 & 0.74 & 0.80 & 0.80 & 0.32\\

        \bottomrule
        \end{tabular}%
        }
        \label{tab:human_align}
    \end{minipage}
    \hfill
    \begin{minipage}[t]{0.39\textwidth}
        \centering
        \caption{Correlation validation of OPD-based VLM on test set.}
        \resizebox{1\textwidth}{!}{%
        \begin{tabular}{c|c|ccc}
        \toprule
            Correlation& VLM & C1 & D1 & D3 \\
        \midrule
            \multirow{2}*{Pearson $\uparrow$}  & Qwen3.5-Omni & 0.72 & 0.90 & 0.87 \\
             ~ & Qwen3-VL-8B  & 0.70 & 0.76 & 0.81\\
    
             \midrule
            
            \multirow{2}*{Spearman $\uparrow$} & Qwen3.5-Omni & 0.80 & 1.00 & 0.78 \\
            ~ & Qwen3-VL-8B  & 0.80 & 0.80 & 0.66 \\

        \bottomrule
        \end{tabular}%
        }
        \label{tab:human_align1}
    \end{minipage}
\end{table*}

In addition, we conducted manual positive and negative sample tests, sensitivity analysis and overall score aggregation analysis. 
The results are presented in Appendix \ref{app:posi_neg} to \ref{app:aggregation}.

\section{Conclusion}

We present \textbf{CutCraft}, a benchmark for evaluating editing-technique execution in multi-shot audio-video generation. CutCraft moves beyond conventional video benchmarks by targeting professional editing constructs, including narrative structure, rhythm, montage, transition design, cinematography, and audio. To this end, we build a strictly curated dataset with structured prompts, editing-aware labels, and fine-grained questions, together with a hierarchical evaluation framework based on shot alignment and evidence-aware scoring. Our experiments reveal a consistent gap between coherence and editing-technique execution: current models can generate visually plausible videos, but remain weak at executing montage logic, transition design, and controlled audio-video relations. We further show that editing-aware decomposition improves several dimensions of discrete control, and that our proposed metrics align closely with expert judgment. We hope CutCraft will support future work on editing-aware generation and evaluation, and help shift video generation from visual plausibility toward genuine editorial control.

\bibliography{iclr2027_conference}
\bibliographystyle{iclr2027_conference}

\newpage
\appendix


\section*{Appendix}

\section{Dataset Details}

\subsection{Data Construction Details}
\label{app:data_construction}

\subsubsection{Meta Prompt}
\label{app:meta}

Our data construction starts from meta-prompt, which is a professionally defined four-tuple consisting of the montage subtype, video content category, visual style, and number of shots. 

\paragraph{Montage Subtypes.}
Specifically, the montage subtypes include \textbf{Narrative Montage}, which presents events clearly according to temporal or causal logic, with the goal of helping the audience understand the story, \textbf{Expressive Montage}, which creates emotional impact and deeper implications through the collision of shots, with the goal of making the audience feel emotion, and \textbf{Intellectual Montage}, which expresses abstract concepts, ideas, or ideology through relationships between images, with the goal of provoking thought in the audience. 

Narrative montage includes \textbf{Sequential Montage}, which follows a single storyline in chronological order; 
\textbf{Parallel Montage}, which presents different storyline threads from different times or spaces in parallel and eventually brings them together; 
\textbf{Crosscut Montage}, which rapidly alternates between multiple storyline threads occurring at the same time but in different locations, where the threads influence each other, often to create tension and suspense; 
\textbf{Repetition Montage}, which repeatedly reintroduces shots with special significance, such as important objects or actions, at key moments in order to shape character portrayal or elevate the theme; 
and \textbf{Dialogue Montage}, which separates questions and answers from the same conversation into different scenes, aiming to compress time or space, omit intermediate processes, and create rapid association between different plotlines. 

Expressive montage includes \textbf{Lyrical Montage}, which inserts empty shots or poetic imagery into narrative sequences in order to intensify mood and atmosphere; 
\textbf{Psychological Montage}, which uses shot combinations to directly present a character’s dreams, memories, hallucinations, or imagination; 
\textbf{Metaphorical Montage}, which conveys meaning implicitly through visual analogy between shots; 
\textbf{Contrast Montage}, which juxtaposes sharply opposed content or form, such as love versus hatred, innocence versus evil, or warm versus cold visual tones, in order to generate strong conflict and reinforce the theme; 
and \textbf{Accumulative Montage}, which rapidly assembles a series of shots with similar properties or related content, accumulating emotion, intensifying atmosphere, and summarizing an overall impression within a short time.

Intellectual montage includes \textbf{Montage of Attractions}, which inserts shots that may appear unrelated to the plot in order to create emotional shock and guide the audience toward a particular idea or attitude; 
\textbf{Reflexive Montage}, which is similar in spirit to montage of attractions, but the metaphorical element already exists within the narrative space, such as a statue or set design; 
and \textbf{Ideological Montage}, which typically reorganizes pre-existing documentary or news footage to argue for or express a particular viewpoint, often carrying documentary and reflective characteristics.

\paragraph{Video Contents.}
The video content categories are defined as follows:
\textbf{Complex action}, which emphasize intricate subject movements, such as martial arts combat, choreographed fighting, weapon-based combat, parkour, street dance battles, ballet, modern dance, basketball, swimming, gymnastics, boxing, skateboarding, rock climbing, surfing, low-altitude skydiving, bungee jumping, wingsuit flying, card shuffling, surgical suturing and magic tricks.
\textbf{Vlog/livestream}, typically human-centered and presented from either a first-person or third-person perspective, including food sharing, campus life sharing, outfit sharing, gaming experience sharing, fitness routines, family life, e-commerce live selling and street live-streaming.
\textbf{Tutorial}, usually object-centered and presented from either a first-person or third-person perspective, such as cooking tutorials, origami, painting, calligraphy, instrument fingering instruction, dance tutorials, vocal training, makeup tutorials, woodworking, welding and agricultural instruction.
\textbf{Stage performance}, which involve complex audio and complicated environments, including solo singing in different styles, choir, band performance, conducting, instrumental solos, concerts, speeches, stand-up comedy, hosting, debates and variety shows.
\textbf{Multi-person dialogue}, characterized by multiple audio sources and complex scenes, such as family dinners, chance encounters, classroom discussions, hospital visits, conversations on the move, news interviews, talk-show panels, elevator small talk, whispering exchanges and casual gossip.
\textbf{Scientific phenomena}, which emphasize scientific plausibility and laboratory settings, such as acid-base reactions, crystallization, combustion, microscopic observation, dissection, electromagnetic induction, fluid mechanics, spring motion, collisions, and astronomical observation.
\textbf{Advertisement}, which often focus on close-up presentation of products, such as sneakers, phones and computers, cars, household appliances, medicine, perfume spray, food and beverages, skincare products, game trailers, sports promotion videos, tourism advertisements and e-commerce product displays.
\textbf{Natural phenomena}, which emphasize scientific plausibility in natural environments, such as waterfalls, solar eclipses, meteor showers, auroras, volcanic eruptions, deep-sea bioluminescence, animal camouflage, forest fires, sandstorms, mudslides, glacier collapse, polar night and monsoon rainfall.
\textbf{Special point-of-view}, which involve uncommon and complex viewing conditions, such as mall monitor, station monitor, exam-room monitor, bus monitor, bank monitor, dashcam footage, bodycam footage, covert journalistic recording, and peephole views.

\paragraph{Visual Styles.}
The visual style categories include \textbf{Naturalistic Realism}, \textbf{Documentary Style}, \textbf{Cinematic Narrative Style}, \textbf{Commercial Advertising Style}, \textbf{Music Video Style}, \textbf{Experimental Style}, \textbf{Sci-fi Futuristic Style}, \textbf{Vintage / Retro Style}, and \textbf{Social Media / Vlog Style}.

\paragraph{Shot numbers.}
The number of shots ranges from \textbf{2} to \textbf{6}. Since our prompts include many complex transition instructions, a maximum of 6 shots (with an average duration of 2.5 seconds per shot) is a relatively appropriate setting. If the shot duration is too short, the editing instructions themselves may not be suitable to be fully executed, which would introduce additional negative interference when evaluating model performance.

\subsubsection{Draft}
\label{app:draft}

Since our final structured prompts contain multiple components, including video content, intra-shot attributes, and cross-shot transitions, generating them directly from the meta prompts in a single step leads to unsatisfactory results. Therefore, after obtaining the filtered meta prompts, we first use an LLM to expand them into drafts. 

The draft generation process is guided by instructions, which include the definitions and intended meanings of the montage subtypes introduced in the previous section, as well as several constraint conditions. 

\begin{tcolorbox}[
  enhanced,breakable,
  colback=gray!4,colframe=black!60,
  arc=2pt,boxrule=0.5pt,
  fonttitle=\bfseries\small,
  title={Draft Generation System Prompt}
]
\scriptsize\ttfamily
You are an expert screenwriter and prompt designer who crafts vivid, physically-grounded
video scene descriptions. For each input META, you produce a rich textual draft that captures the
creative vision, story arc, and physical/audio-video details of a 15-second multi-shot video.\\

Your output is a TEXTUAL DRAFT — not a final structured prompt. Focus on:\\
  1. Inventing a compelling, specific scenario that embodies the seed's four constraints\\
  2. Writing vivid prose that describes what happens across the entire 15-second video\\
  3. Breaking down the video into the specified number of shots with clear per-shot descriptions\\
  4. Embedding at least one non-trivial physically-grounded audio-video coupling phenomenon\\

Constraints from the META:\\
  (1) Performance method [A] — controls the editing logic and narrative flow\\
  (2) Video content [B]     — defines the subject / topic\\
  (3) Visual style [C]      — defines the look-and-feel\\
  (4) Number of shots [D]   — MUST be respected exactly\\

A-category performance method DEFINITIONS (you MUST design the scene's narrative flow and
shot transitions to embody the assigned method):
\{MONTAGE\_DEFINITIONS\} \\

\{FORMAT\_REQUIREMENTS\} \\

Hard requirements:\\
- "results" array length MUST equal the number of input METAs, in the SAME ORDER.\\
- Each "id" MUST match the SEED INDEX of the corresponding seed.\\
- "number\_of\_shots" MUST equal the seed's shots value; "shot\_descriptions" array length MUST match.\\
- "time\_range" must tile [00.00s, 15.00s] with no gaps/overlaps; last shot ends at 15.00s.\\
- "main\_description" must be in English, rich and specific (not generic placeholder text).\\
- Each "shot\_descriptions[].description" must be a substantial paragraph (at least 50 words) in English.\\
- ONE SHOT = ONE CONTINUOUS TAKE: Each shot description MUST describe ONLY ONE unbroken continuous scene from ONE camera position/location. STRICTLY FORBIDDEN within a single shot:\\
  • "cut to", "switch to", "intercut", "then we see", "meanwhile", "alternating between"\\
  • Describing two or more distinct locations, subjects, or viewpoints\\
  • Any language implying a transition or scene change within the shot\\
  For montage types requiring alternation:\\
  the alternation MUST be achieved by assigning different storylines to SEPARATE shots,
  NOT by cramming multiple storylines into one shot description.\\
- Output ONLY valid JSON; no markdown, no code-fence, no comments, no extra text.

\end{tcolorbox}

In the above prompt, {\ttfamily \{MONTAGE\_DEFINITIONS\}}  refers to the previously defined introduction to montage types, and {\ttfamily\{FORMAT\_REQUIREMENTS\}} specifies the mandatory formatting constraints, as detailed below:

\begin{tcolorbox}[
  enhanced,breakable,
  colback=gray!4,colframe=black!60,
  arc=2pt,boxrule=0.5pt,
  fonttitle=\bfseries\small,
  title={Draft Format Requirements}
]
\scriptsize\ttfamily
\{\\
\hspace*{1em}"results": [\\
\hspace*{2em}\{\\
\hspace*{3em}"id": \textless int --- MUST equal the META INDEX\textgreater,\\
\hspace*{3em}"main\_description": "\textless A rich paragraph (200-400 words) in ENGLISH describing the entire 15-second video: setting, characters, lighting, mood, physical phenomena, sound design, the A-method narrative logic, and the style atmosphere. Must include the specific physics/audio-video coupling element. Must be horizontally framed (16:9).\textgreater",\\
\hspace*{3em}"number\_of\_shots": \textless int --- MUST equal META.shots\textgreater,\\
\hspace*{3em}"shot\_descriptions": [\\
\hspace*{4em}\{\\
\hspace*{5em}"shot\_id": 1,\\
\hspace*{5em}"time\_range": "00.00-XX.XXs",\\
\hspace*{5em}"description": "\textless Detailed ENGLISH description of this shot: what is visible, what is audible, what physical phenomenon occurs, how it connects to the previous/next shot in the context of the A-method montage logic. Include subject, action, environment, lighting, sound, and any physics detail.\textgreater",\\
\hspace*{4em}...\\
\}
\end{tcolorbox}

\subsubsection{Structured Prompt}

After obtaining the draft, we further inject finer-grained cinematographic and editing knowledge into the LLM, and ask it to expand the draft into a structured prompt. Specifically, the shot-level details include shot scale, camera angle, camera movement, focus, depth of field, and other intra-shot attributes. The transition-level descriptions include camera-related transition dimensions, visual-effect dimensions, audio-video relations, temporal overlap, and detailed transition instructions.

\paragraph{Intra-shot Cinematography Vocabulary.}
To ensure consistency in structured prompt generation, we adopt a standardized vocabulary for describing intra-shot cinematographic attributes, including shot scale, camera angle, camera movement, optical motion, and depth of field / focal length.

\textbf{Shot Scale} specifies the visual distance between the camera and the subject. \textbf{Extreme Long Shot (ELS)} emphasizes a vast landscape in which the subject is barely visible. \textbf{Long Shot (LS) / Wide Shot} retains strong environmental context while keeping the subject relatively small in the frame. \textbf{Full Shot (FS)} shows the subject’s full body. \textbf{Medium Shot (MS)} typically frames the subject from the waist up. \textbf{Medium Close-Up (MCU)} frames the subject from the chest or bust upward. \textbf{Close-Up (CU)} fills the frame with a face or another important object. \textbf{Extreme Close-Up (ECU)} isolates a very small detail, such as an eye, a finger, or a surface texture.

\textbf{Camera Angle} describes the viewpoint from which the subject is filmed. \textbf{Eye-Level Shot} provides a neutral and natural perspective. \textbf{High-Angle Shot} positions the camera above the subject, often suggesting vulnerability or offering an overview. \textbf{Low-Angle Shot} places the camera below the subject, often implying power or grandeur. \textbf{Bird’s-Eye View (Overhead)} looks straight downward, creating an abstracted geometric impression. \textbf{Worm’s-Eye View (Ground)} uses an extreme upward angle, producing strong dramatic foreshortening. \textbf{Dutch Angle (Canted Tilt)} tilts the frame in order to create tension or unease. \textbf{Over-the-Shoulder (OTS)} frames one character through the shoulder line of another. \textbf{POV Shot (First-Person)} presents the scene directly from a character’s perspective.

\textbf{Camera Movement} refers to the physical displacement or rotation of the camera body. \textbf{Static (Locked-Off)} means that the camera remains fixed. \textbf{Pan} rotates the camera horizontally around a fixed axis. \textbf{Tilt} rotates the camera vertically around a fixed axis. \textbf{Roll} rotates the camera around the lens axis. \textbf{Dolly-In / Push-In} physically moves the camera toward the subject. \textbf{Dolly-Out / Pull-Out} physically moves the camera away from the subject. \textbf{Tracking Shot (Follow)} moves the camera together with the subject along a path. \textbf{Truck (Lateral Tracking)} moves the camera sideways, perpendicular to the lens axis. \textbf{Arc Shot (Orbit)} circles around the subject. \textbf{Crane Shot (Boom / Pedestal)} lifts or lowers the camera using a crane or pedestal. \textbf{Handheld Shot} is operated by hand and therefore carries a more unstable, visceral, or documentary-like quality. \textbf{Steadicam Shot} uses stabilization to achieve movement that is smooth while still retaining an organic feel.

\textbf{Optical / Hybrid Motion} is distinguished from physical camera movement because it is produced optically or through a hybrid combination. \textbf{Zoom-In} increases focal length so that the subject appears larger without moving the camera body. \textbf{Zoom-Out} decreases focal length so that the subject appears smaller. \textbf{Dolly Zoom (Vertigo / Hitchcock Zoom)} combines a physical dolly motion with an opposing zoom, creating a surreal distortion of spatial perception. \textbf{None} indicates that no optical motion is used.

\textbf{Depth of Field and Focal Length} describe how focus and lens properties shape spatial appearance. \textbf{Shallow Depth of Field (Shallow DoF)} keeps the subject sharp while blurring the background, often through telephoto settings or wide apertures. \textbf{Deep Focus} keeps both foreground and background in focus simultaneously. \textbf{Wide-Angle Lens} expands the field of view and may introduce mild edge distortion. \textbf{Standard Lens (Normal)} provides a natural and relatively undistorted perspective. \textbf{Telephoto Lens} compresses spatial depth and isolates distant subjects. \textbf{Macro Lens} enables extreme close-up magnification of very small subjects. \textbf{Tilt-Shift Lens} manipulates the focus plane selectively and can create a miniature-like visual effect.

\paragraph{Inter-shot Transition Vocabulary.}
For cross-shot editing, we also define a standardized vocabulary that covers cinematographic transition type, optical or effects-based transition, and audio--visual relationship.

\textbf{Cinematographic Transition Type} describes how the cut is motivated within the visual or narrative logic. \textbf{Graphic Match} emphasizes visual continuity across shots by preserving similar subject matter, shape, color, or composition. \textbf{Match on Action} links two consecutive shots through the same or continuous action, even when the camera viewpoint changes. \textbf{Sound Match (Audio Match)} creates continuity by overlapping or matching similar audio elements across the cut. \textbf{Contrast Cut} deliberately places adjacent shots in sharp contrast in terms of scale, movement, or tonality, producing a strong visual or thematic break. \textbf{Extreme Scale Cut (Polar-Scale Cut)} juxtaposes shots with maximally different scales, such as from an extreme long shot to an extreme close-up, creating a striking rhythmic contrast. \textbf{Whip-Pan Transition (Camera-Movement Transition)} uses rapid camera movement to bridge otherwise separate scenes. \textbf{POV Cut} moves from the person who is looking to the object or scene being seen. \textbf{Exit-and-Entry Transition} connects two shots by having a subject or moving object exit the frame in one shot and another enter the frame in the next. \textbf{Wipe-By (Foreground-Occlusion Transition)} uses a foreground object to temporarily block the frame and conceal the cut. \textbf{Empty Shot Transition (Cutaway / Insert)} inserts a character-free shot of a landscape or object in order to convey mood, reflection, or psychological state. \textbf{Logical Cut (Causal Transition)} is driven by narrative causality, where the following shot answers, responds to, or fulfills the previous one.

\textbf{Optical / Effects Transition} specifies the visual effect at the moment of transition. \textbf{Hard Cut} changes instantly from one shot to another without any optical effect. \textbf{Dissolve} gradually overlaps the outgoing and incoming shots. \textbf{Wipe} replaces one image with another through a directional movement across the frame. \textbf{Flash-to-Black} inserts a brief full-black frame between shots. \textbf{Flash-to-White} inserts a brief full-white frame between shots.

\textbf{Audio-video Relationship} describes how sound and image align across the cut. \textbf{J-Cut} introduces the audio of the incoming shot before its image appears. \textbf{L-Cut} lets the audio of the outgoing shot continue after the image has already changed. \textbf{Straight Cut} cuts both image and sound at the same frame.

The system prompt of prompt generation is presented as follows:

\begin{tcolorbox}[
  enhanced,breakable,
  colback=gray!4,colframe=black!60,
  arc=2pt,boxrule=0.5pt,
  fonttitle=\bfseries\small,
  title={System Prompt of Structed Prompt Generation}
]
\scriptsize\ttfamily
You are an expert cinematographer and prompt engineer who converts textual video scene\\
drafts into precisely structured video production prompts.\\[3pt]

You receive DRAFT descriptions (produced in an earlier step) that contain:\\
\hspace*{1em}- A main description of the 15-second video scene\\
\hspace*{1em}- The number of shots and per-shot textual descriptions\\
\hspace*{1em}- The assigned montage type, content topic, and visual style\\[3pt]

Your job is to:\\
\hspace*{1em}1. Preserve ALL creative content from the draft (story, physics phenomena, characters, setting)\\
\hspace*{1em}2. Add precise cinematographic details: shot scale, angle, camera motion, optical motion, DoF\\
\hspace*{1em}3. Design appropriate inter-shot transitions that are LOGICALLY consistent with the content\\

\{VOCABULARY\} \\

\{FORMAT\_REQUIREMENTS\} \\

\end{tcolorbox}

Here, {\ttfamily\{VOCABULARY\} }refers to the terminology definitions introduced above. The final output {\ttfamily\{FORMAT\_REQUIREMENTS\} }are as follows:

\begin{tcolorbox}[
  enhanced,breakable,
  colback=gray!4,colframe=black!60,
  arc=2pt,boxrule=0.5pt,
  fonttitle=\bfseries\small,
  title={Structured Prompt Generation System Prompt}
]
\scriptsize\ttfamily
"results": [\\
\hspace*{1em}\{\\
\hspace*{2em}"id": \textless int --- MUST equal the DRAFT's id\textgreater,\\
\hspace*{2em}"scene\_en": \{\\
\hspace*{3em}"id": \textless same int\textgreater,\\
\hspace*{3em}"source\_seed": "\textless copy source\_seed\_en from the draft\textgreater",\\
\hspace*{3em}"title": "\textless concise English title\textgreater",\\
\hspace*{3em}"overall\_description\_prompt": "\textless rich English paragraph: subject, setting, lighting, mood, sound design, EXPLICIT mention of the physical / audio-video coupling phenomenon, EXPLICIT note that the video is 15s and 16:9 landscape\textgreater",\\
\hspace*{3em}"global\_editing\_style": "\textless short paragraph; name the assigned montage type; explain how its specific logic manifests across shot structure and transition design\textgreater",\\
\hspace*{3em}"number\_of\_shots": \textless int --- MUST equal draft's number\_of\_shots\textgreater,\\
\hspace*{3em}"shots": [\\
\hspace*{4em}\{\\
\hspace*{5em}"shot\_id": 1,\\
\hspace*{5em}"description\_prompt": "Shot 1 [00.00-XX.XXs]: \textless visual content, lighting, sound, motion, physics details; include explicit horizontal-framing language\textgreater",\\
\hspace*{5em}"camera": \{\\
\hspace*{6em}"shot\_scale": "\textless ELS | LS (wide-shot) | FS (full shot) | MS | MCU | CU (close-up) | ECU (extreme close-up)\textgreater",\\
\hspace*{6em}"angle": "\textless eye-level | high-angle | low-angle | bird's-eye (overhead) | worm's-eye (ground-level) | dutch angle (canted tilt) | OTS (over-the-shoulder) | POV (first-person)\textgreater",\\
\hspace*{6em}"camera\_motion": "\textless static (locked-off) | pan | tilt | roll | dolly-in (push-in) | dolly-out (pull-out) | tracking (follow) | truck (lateral) | arc (orbit) | crane (boom/pedestal) | handheld | steadicam\textgreater",\\
\hspace*{6em}"optical\_motion": "\textless zoom-in | zoom-out | dolly zoom (vertigo/Hitchcock zoom) | none\textgreater",\\
\hspace*{6em}"depth\_of\_field": "\textless shallow DoF | deep focus | wide-angle lens | standard lens | telephoto | macro | tilt-shift | unspecified\textgreater"\\
\hspace*{5em}\},\\
\hspace*{5em}"transition\_to\_next": \{\\
\hspace*{6em}"cinematographic\_type": "\textless graphic match | match on action | sound match (audio match) | contrast cut | extreme scale cut (polar-scale cut) | whip-pan transition | POV cut | exit-and-entry transition | wipe-by (foreground-occlusion) | empty shot transition (cutaway/insert) | logical cut (causal transition) | none\textgreater",\\
\hspace*{6em}"optical\_effect": "\textless hard cut (straight cut) | dissolve (cross-dissolve) | wipe | flash-to-black | flash-to-white\textgreater",\\
\hspace*{6em}"audio\_visual\_relation": "\textless J-cut | L-cut | straight cut (sync cut)\textgreater",\\
\hspace*{6em}"timing\_offset\_seconds": \textless float --- audio pre-roll / overlap window in seconds, e.g. 0.8\textgreater,\\
\hspace*{6em}"transition\_duration\_seconds": \textless float --- optical effect duration in seconds, e.g. 0.5\textgreater,\\
\hspace*{6em}"description": "\textless narrative description: which cinematographic device, which optical effect, and audio-video relationship\textgreater"\\
\hspace*{5em}\}\\
\hspace*{4em}\},\\
\hspace*{4em}...,\\
\hspace*{4em}\{\\
\hspace*{5em}"shot\_id": \textless last\textgreater,\\
\hspace*{5em}"description\_prompt": "...[XX.XX-15.00s]...",\\
\hspace*{5em}"camera": \{\\
\hspace*{6em}"shot\_scale": "...", "angle": "...", "camera\_motion": "...",\\
\hspace*{6em}"optical\_motion": "...", "depth\_of\_field": "..."\\
\hspace*{5em}\}...\\
\end{tcolorbox}

Notably, the description of the final shot does not include transition-related fields.

\subsubsection{Question Bank}

Based on the generated structured prompts, we invited experts with professional knowledge to build the corresponding question bank required for evaluation. 
The evaluation dimensions include A2 Event Execution Alignment, A3 Causal-Chain Plausibility, C1 Montage Type, E3 Cross-Shot Physical Consistency, E1 Camera-Parameter Combination, and D1 Transition Type Recognition. 

\begin{tcolorbox}[
  enhanced,breakable,
  colback=blue!4,colframe=blue!50!black,
  arc=2pt,boxrule=0.5pt,
  fonttitle=\bfseries\small,
  title={Question Bank Construction Guideline for Experts}
]
\scriptsize

Human experts were asked to construct \textbf{exactly six evaluation questions} for each video based on its structured prompt description. The question bank should cover the six target dimensions in the following order: \textbf{A2, A3, C1, E3, E1, and D1}.

\paragraph{General rules for option design.}
For all single-choice questions, annotators were instructed to ensure that:
\begin{enumerate}
    \item The lengths of options A--D are balanced. The difference between the longest and shortest option should not exceed three words.
    \item For A2, A3, E1, and D1, each question contains five options: A--D plus \textbf{E. None of the above descriptions is correct}.
    \item For C1, each question contains \textbf{exactly four options} (A--D), with no option E.
    \item For A2, each option must follow the format \texttt{event1 $\rightarrow$ event2 $\rightarrow \cdots \rightarrow$ eventN}, where \(N\) is the number of shots in the video.
    \item For E1, all four options must follow the same format: \texttt{XX -- XX -- XX -- XX}.
    \item For D1, all four options must be standard English names of cinematographic transition types in a unified style.
    \item Annotators should avoid making the correct answer obviously longer, more detailed, or linguistically distinct from the distractors.
\end{enumerate}

\paragraph{Question design requirements.}
\begin{enumerate}
    \item \textbf{A2 Event Execution Alignment (single choice).}  
    Ask: \textit{``Which option correctly describes the event order of the shots in the video?''}  
    Each option should describe the event sequence of all shots using the format \texttt{event1 $\rightarrow$ event2 $\rightarrow \cdots \rightarrow$ eventN}.  
    The correct option should list the core event of each shot in order, while distractors may replace, shuffle, or invent events.

    \item \textbf{A3 Causal-Chain Plausibility (single choice).}  
    Ask: \textit{``What is the editing logic between shot X and shot Y?''}  
    Annotators should provide four plausible English options, each describing a possible causal or narrative connection between the two shots. The correct option should summarize the true editing logic.

    \item \textbf{C1 Montage Type (single choice).}  
    Ask: \textit{``Which montage sub-type best describes the editing structure of this video?''}  
    Exactly four options (A--D) should be provided, all expressed as montage sub-type names. The correct answer and distractors should come from the same major category whenever possible:
    \begin{itemize}
        \item Narrative: Sequential, Parallel, Crosscut, Repetition, Dialogue
        \item Expressive: Lyrical, Psychological, Metaphorical, Contrast, Accumulative
        \item Intellectual: Montage of Attractions, Reflexive, Ideological
    \end{itemize}
    If the correct answer belongs to the Intellectual category, annotators should use all three Intellectual sub-types and add one Narrative sub-type as the fourth option.  
    The following pairing rules must be respected:
    \begin{itemize}
        \item Parallel Montage and Crosscut Montage must appear together.
        \item Contrast Montage and Accumulative Montage must appear together.
        \item Metaphorical Montage and Psychological Montage must appear together.
    \end{itemize}

    \item \textbf{E3 Cross-Shot Physical Consistency (true/false).}  
    Ask: \textit{``Does the phenomenon XXX in the video satisfy physical laws?''}  
    Here, XXX should refer to a physical consistency phenomenon identified from the prompt, such as object shape consistency, appearance consistency, lighting continuity, motion continuity, or stable spatial relations across shots. The phenomenon must involve at least two shots.  
    Only two options are allowed:  
    \texttt{A. Yes}  
    \texttt{B. No}

    \item \textbf{E1 Camera-Parameter Combination (single choice).}  
    Select one shot at random and ask: \textit{``What is the shot-scale -- angle -- camera-motion -- depth-of-field combination for shot X?''}  
    All four options should use the unified format \texttt{XX -- XX -- XX -- XX}.

    \item \textbf{D1 Transition Type Recognition (single choice).}  
    Select one transition at random and ask: \textit{``What is the cinematographic transition type between shot X and shot Y?''}  
    The four options must be chosen exclusively from the following cinematographic transition types:
    \begin{enumerate}
        \item Graphic Match / Similar Visual 
        \item Match on Action / Action Match
        \item Sound Match / Audio Match
        \item Contrast Cut
        \item Extreme Scale Cut / Polar-Scale Cut
        \item Whip-Pan Transition / Camera-Movement Transition
        \item Point-of-View Cut / POV Cut
        \item Exit-and-Entry Transition / Walk-Out Walk-In Cut
        \item Wipe-By Transition / Foreground-Occlusion Transition
        \item Cutaway to Scenery / Empty Shot Transition / Insert Shot
        \item Logical Cut / Causal Transition
    \end{enumerate}
\end{enumerate}


\end{tcolorbox}

\subsubsection{Event Coherence Label}

The \textit{Event Coherence Label} is an important cue in our dataset and is closely related to montage type. We divide montage types into two categories. The first is \textit{Montage with Continuous Chain}, which means that the development of the event chain remains continuous throughout the video. This category includes {sequential}, {repetition}, {dialogue}, {psychological}, {accumulative}, and {reflexive} montage. The second is \textit{Montage with Discontinuous Chain}, which means that the video involves shifts in the event chain during narration, including changes in time or space. This category includes {parallel}, {crosscut}, {lyrical}, {metaphorical}, {contrastive}, {attractions}, and {ideological} montage.

For samples belonging to \textit{Montage with Continuous Chain}, all shots are assigned the label 0. For samples belonging to \textit{Montage with Discontinuous Chain}, the labels are determined according to the content organization of each prompt and its montage type. Shots that belong to the same event chain are assigned the same numeric label, and the absolute values of the labels are ordered from small to large according to their first appearance. For example, a five-shot sequential montage is labeled as \([0,0,0,0,0]\), while a five-shot crosscut montage may be labeled as \([0,1,0,1,0]\).

\subsection{Data Analysis Details}
\label{app:data_analysis}

We provide the dataset analysis from three perspectives: the video level, the transition level, and the shot level.

\paragraph{Video-level analysis.}
At the video level, the dataset contains a total of 295 videos. In terms of montage composition, \textit{Narrative Montage} accounts for $43.1\%$ of the dataset, \textit{Expressive Montage} accounts for $35.9\%$, and \textit{Intellectual Montage} accounts for $21.0\%$, indicating that the dataset not only fully accounts for data diversity, but also preserves a distribution consistent with that of everyday narrative logic. 
Among the fine-grained montage subtypes, \textit{Sequential Montage} is the most frequent, with 31 samples ($10.5\%$), followed by \textit{Parallel Montage} with 26 samples ($8.8\%$). \textit{Crosscut Montage} and \textit{Repetition Montage} each contain 24 samples ($8.1\%$), while \textit{Dialogue Montage}, \textit{Psychological Montage}, \textit{Contrast Montage}, and \textit{Ideological Montage} each account for $7.5\%$. 
In terms of content, the dataset spans a diverse set of video scenarios, including vlog / livestream, complex action, advertisement, stage performance, tutorial, scientific phenomena, multi-person dialogue, natural phenomena, and special POV content, showing that montage forms are distributed across heterogeneous semantic domains. In terms of visual style, \textit{Vintage / Retro Style} and \textit{Documentary Style} are the most common, accounting for $14.2\%$ each, followed by \textit{Naturalistic Realism} at $13.2\%$, while \textit{Experimental Style} is the least frequent at $7.8\%$. The number of shots per video ranges from 2 to 6, with 5-shot videos being the most common at $24.1\%$, followed by 4-shot videos at $23.1\%$, suggesting that medium-length multi-shot structures are the dominant format in the dataset.

\paragraph{Transition-level analysis.}
At the transition level, the dataset contains a total of 933 transitions. For cinematographic transition types, \textit{Graphic Match} is the most frequent, accounting for $13.6\%$, followed by \textit{Match on Action} and \textit{Logical Cut}, both at $11.9\%$. Other common categories include \textit{Sound Match} ($10.5\%$), \textit{POV Cut} ($9.3\%$), and \textit{Extreme Scale Cut} ($8.8\%$), indicating that the dataset covers a broad range of transition logic. In terms of optical transition effects, \textit{Dissolve} is the most common, accounting for $38.0\%$ of all transitions, followed by \textit{Hard Cut} at $32.3\%$. \textit{Flash-to-White}, \textit{Wipe}, and \textit{Flash-to-Black} account for $13.0\%$, $12.2\%$, and $4.5\%$, respectively. For audio-video relations, \textit{J-Cut} is the dominant pattern at $40.3\%$, followed by \textit{L-Cut} at $34.4\%$ and \textit{Straight Cut} at $25.3\%$. Overall, the transition statistics suggest that the dataset not only covers diverse visual editing logic, but also places strong emphasis on audio-led and audio-overlapping transition design.

\paragraph{Shot-level analysis.}
At the shot level, the dataset contains a total of 1228 shots. Regarding shot scale, \textit{Close-Up} is the most frequent at $24.9\%$, followed closely by \textit{Medium Shot} at $24.7\%$ and \textit{Long Shot} at $21.9\%$, while \textit{Extreme Long Shot} is relatively rare at $2.7\%$. This indicates that the dataset emphasizes subject-centered framing while retaining sufficient environmental coverage. In terms of camera angle, \textit{Eye-Level Shot} dominates at $35.0\%$, followed by \textit{Low-Angle Shot} at $19.9\%$ and \textit{High-Angle Shot} at $19.2\%$; more specialized perspectives such as \textit{OTS}, \textit{Bird's-Eye View}, \textit{Worm's-Eye View}, \textit{Dutch Angle}, and \textit{POV} each occupy smaller but meaningful shares. For camera motion, \textit{Static} is the most common at $17.1\%$, but a wide range of dynamic motions are also well represented, including \textit{Dolly-In} ($13.7\%$), \textit{Truck} ($10.9\%$), \textit{Arc} ($9.8\%$), \textit{Handheld} ($9.4\%$), and \textit{Crane} ($9.0\%$). In optical motion, the majority of shots use no optical manipulation, accounting for $43.1\%$, while \textit{Zoom-In} and \textit{Zoom-Out} account for $25.1\%$ and $19.5\%$, respectively, and \textit{Dolly Zoom} appears in $12.3\%$ of shots. 

\begin{figure}[h!]
    \centering
    \includegraphics[width=1\linewidth]{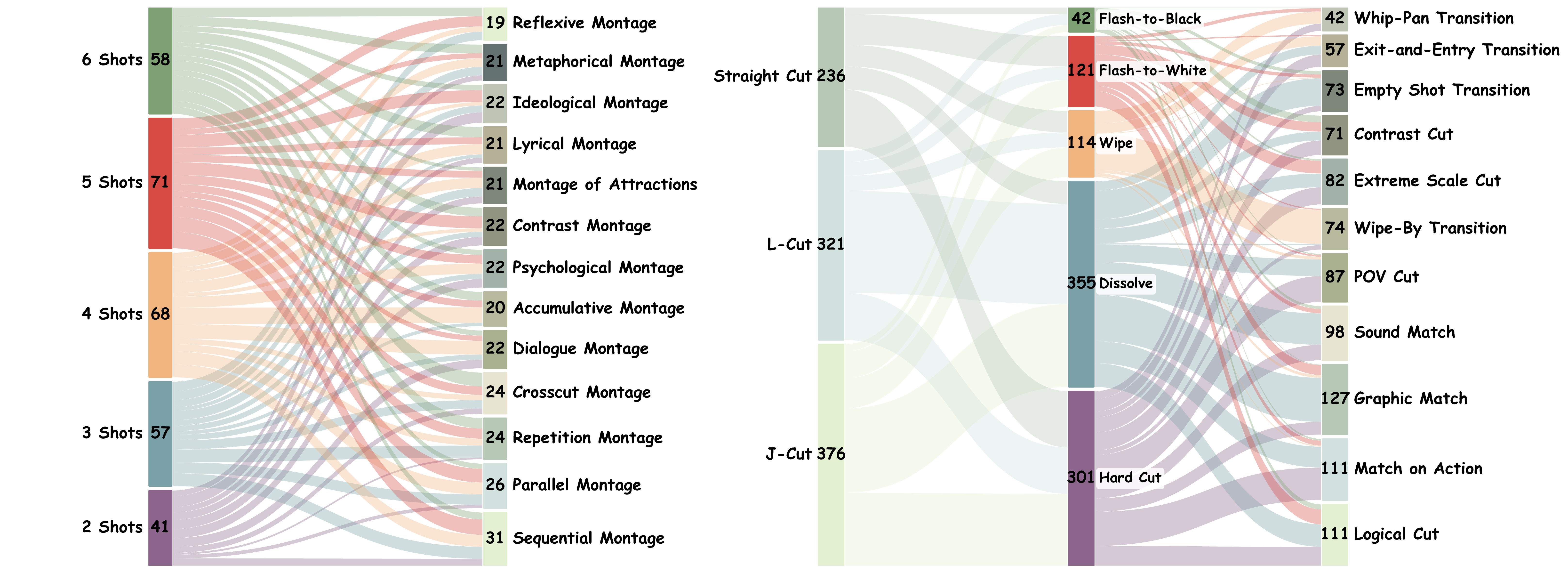}
    \caption{Analysis of shot count and montage type at the video-sample level, and three transition-level dimensions.}
    \label{app:fig_data_sankey}
\end{figure}

\section{Metrics Details}

\subsection{Metrics Design and Evaluation}
\label{app:metrics_design}

Our metric suite is implemented under a hierarchical evaluation framework with a shared shot-alignment pre-processing layer and three evidence pathways: \textbf{Path-I} (\emph{direct expert-model scoring}), \textbf{Path-II} (\emph{expert evidence with VLM arbitration}), and \textbf{Path-III} (\emph{VLM-based question answering and judgment}). Before computing any metric, we first apply \textbf{TransNetV2}~\citep{soucek2024transnet} to obtain raw shot boundaries and then use a VLM-based shot-alignment module to align generated segments with the prompt-defined ground-truth shot plan. The aligned result marks each intended shot as \texttt{matched}, \texttt{merged}, or \texttt{missing}, and is shared by downstream metrics. In what follows, we describe the computation of all metrics in the order of groups \textbf{A--F}, and explicitly indicate which evaluation pathway each metric uses.

\paragraph{A. Narrative execution.}
The A group evaluates whether the generated video realizes the intended shot-structured narrative plan.

\textbf{A1 (shot-count accuracy)} uses \textbf{Path-I}. We directly compare the number of shots detected by TransNetV2~\citep{soucek2024transnet}, denoted by $n_{\mathrm{pred}}$, with the ground-truth number of shots $n_{\mathrm{gt}}$. The score is defined as
\[
A1=\mathrm{clip}_{[0,1]}\!\left(1-\frac{|n_{\mathrm{pred}}-n_{\mathrm{gt}}|}{\max(n_{\mathrm{gt}},1)}\right).
\]
This dimension evaluates only the actual visual shot segmentation and does not apply VLM-based shot alignment. The reason is that, after alignment, the shots are effectively evaluated at the level of the event chain. Therefore, the A1 dimension is designed solely to assess whether the model has realized the intended shot breakdown, thereby distinguishing it from the subsequent dimensions.

\textbf{A2 (event execution alignment)} uses \textbf{Path-III}. It is evaluated in three steps. First, given a multiple-choice event-chain question constructed from the structured prompt, a VLM selects the event order that best matches the generated video. If the selected chain is incorrect, the score is set to zero. 
Here, A2 evaluates whether the overall event sequence is correct and narratively complete, whereas A3 focuses on the logical connection between adjacent shots. For example, a model may correctly generate a character shot followed by a scenery shot, but a stronger model will also create a smooth transition between them, such as having the character turn their gaze at the end of the first shot before cutting to the second to form a POV cut.

Second, for each intended event, we consult the shot-alignment result to determine whether the corresponding shot is present. Missing shots receive zero. Third, for aligned and present shots, we ask the VLM to judge the fidelity of event execution using a three-level scale: \emph{complete} ($1.0$), \emph{partial} ($0.5$), and \emph{failed} ($0.1$). 
It should be noted that the \emph{failed} score of 0.1 indicates that the shot has been generated, but its content is entirely unrelated to the prompt. This is distinct from a score of 0, which corresponds to a shot that is missing altogether.

The final score is the average over all events:
\[
A2=\frac{1}{N}\sum_{i=1}^{N} s_i,
\]
where $N$ is the number of prompt-defined events and $s_i\in\{1.0,0.5,0.1,0\}$ already incorporates shot existence.

\textbf{A3 (causal-chain plausibility)} uses \textbf{Path-III}. It combines two sub-scores. The first sub-score is a VLM-based multiple-choice question asking for the editing logic between relevant shots; it is scored as binary correctness:
\[
A3_{\mathrm{mc}} \in \{0,1\}.
\]
The second sub-score is a \emph{counterfactual order-sensitivity test}. Using the aligned shots, we group shots by \texttt{event\_coherence\_label}, then construct two temporary videos for each event chain: one in the original order and one in reversed shot order, while keeping each shot internally forward in time. A VLM then scores the plausibility of the original order and reversed order as $C_{\mathrm{orig}}$ and $C_{\mathrm{rev}}$. For a given chain, we compute
\[
\mathrm{chain\_score}
=
C_{\mathrm{orig}}
\cdot
\left(
\lambda_0 + (1-\lambda_0)\,
\mathrm{clip}_{[0,1]}(C_{\mathrm{orig}}-C_{\mathrm{rev}})
\right),
\]
where $\lambda_0=0.3$ is an order-prior constant. We average such scores within chains to obtain an intra-chain score, and compute an analogous inter-chain score for montage-level structure. For non-directional montage types, such as lyrical, contrastive, attractions, metaphorical, and reflexive montage, the inter-chain component does not use order reversal and directly takes the original plausibility score. The final counterfactual score is
\[
A3_{\mathrm{cf}}=\frac{A3_{\mathrm{intra}}+A3_{\mathrm{inter}}}{2},
\]
with standard fallback to the available component if one side is missing. The overall A3 score is then
\[
A3=\frac{A3_{\mathrm{mc}}+A3_{\mathrm{cf}}}{2},
\]

\textbf{A4 (style consistency)} uses \textbf{Path-I}. It combines a target-style matching term and a cross-shot style consistency term. The first term is computed by a CLIP-based zero-shot style classifier~\citep{radford2021learning,cherti2023reproducible}, which compares sampled video frames against a fixed set of predefined style text labels. Let $A4_{\mathrm{clip}}$ denote the negative exponential rank of the target style among all style labels. 
The second term is based on \textbf{DINOv2}~\citep{oquab2023dinov2}. We regroup aligned shots according to their \texttt{event\_coherence\_label}, concatenate the shots within each group, and compute the mean cosine similarity between adjacent frame embeddings. 
According to our previous setup, shots with the same \texttt{event\_coherence\_label} belong to the same event chain and should share a consistent overall style. 
This gives a label-level style-consistency score, which is then weighted by the proportion of prompt shots belonging to that label:
\[
A4_{\mathrm{dino}}=\sum_{k} w_k\,\bar{\cos}_k,
\]
where $w_k$ is the prompt-shot ratio of label $k$. The final A4 score is
\[
A4=\frac{A4_{\mathrm{clip}}+A4_{\mathrm{dino}}}{2}.
\]

\paragraph{B. Rhythm and pacing.}
The B group evaluates temporal organization and audio-supported editing rhythm.

\textbf{B1 (shot-duration accuracy)} uses \textbf{Path-I}. Given aligned predicted shot durations $d_i^{\mathrm{pred}}$ and ground-truth durations $d_i^{\mathrm{gt}}$, we compute the normalized mean absolute duration error and convert it into a bounded score:
\[
B1=\mathrm{clip}_{[0,1]}
\left(
1-\frac{1}{M}\sum_{i=1}^{M}\frac{|d_i^{\mathrm{pred}}-d_i^{\mathrm{gt}}|}{d_i^{\mathrm{gt}}}
\right),
\]
where $M$ is the number of evaluable aligned shots. When alignment is unavailable, durations are compared sequentially using the raw TransNetV2~\citep{soucek2024transnet} cuts.
Notably, this dimension mainly evaluates the accuracy of generated shot duration and belongs to the shot-structure level. To avoid double penalization, missing shots are not additionally penalized here, as they have already been accounted for in other shot-structure-related dimensions (A1).

\textbf{B2 (beat synchronization)} uses \textbf{Path-I}. It combines music-motion correlation and transition sound-effect matching. First, we separate the audio into vocals, drums, bass, and other stems using \textbf{HT-Demucs}~\citep{rouard2023hybrid}. If the vocals dominate the non-vocal stems or the music-track RMS is too low, the clip is judged to contain no valid background music. For music-bearing clips, we compute the Pearson correlation between the music-energy envelope and the video motion-energy curve. Motion energy is derived from \textbf{RAFT} optical flow~\citep{teed2020raft}, and music energy is obtained from the non-vocal stems using \texttt{librosa}~\citep{mcfee2015librosa}. The resulting sub-score is
\[
B2_{\mathrm{corr}}=\mathrm{clip}_{[0,1]}\left(\frac{\rho+1}{2}\right),
\]
where $\rho$ is the Pearson correlation coefficient. 

The second sub-score evaluates whether transition sound effects are compatible with the intended optical transition type. Around each cut, we compute local RMS impact ratios within temporal windows near the cut. These impact ratios are then mapped to heuristic scores according to the expected acoustic behavior of the target optical effect: for example, flash transitions prefer stronger impact, while dissolves prefer smoother energy changes. Averaging over transitions gives $B2_{\mathrm{sfx}}$. The final score is
\[
B2=\mathrm{clip}_{[0,1]}\left(\frac{B2_{\mathrm{corr}}+B2_{\mathrm{sfx}}}{2}\right).
\]

\textbf{B3 (rhythm--mood matching)} uses \textbf{Path-II}. It combines a VLM-based rhythm judgment with an expert-model emotion-matching score. The first sub-score, $B3_{\mathrm{rhythm}}$, is produced by a VLM that judges whether the editing pace and sound rhythm fit the scene, using evidence such as cut rate and temporal activity.

The second sub-score, $B3_{\mathrm{mood}}$, is based on \textbf{PANNs}~\citep{kong2020panns} for audio event and affective tagging. From the prompt we infer the expected emotional tone, then compare it with the predicted music-emotion tags. If $\mathrm{match\_ratio}$ and $\mathrm{conflict\_ratio}$ denote the fractions of matching and conflicting affective cues, we define
\[
B3_{\mathrm{mood}}=\mathrm{match\_ratio}-0.5\cdot \mathrm{conflict\_ratio}.
\]
The final B3 score is
\[
B3=\frac{B3_{\mathrm{rhythm}}+B3_{\mathrm{mood}}}{2}.
\]

\paragraph{C. Montage execution.}
The C group targets montage, which is inherently an across-shot editorial construct.

\textbf{C1 (montage type)} combines \textbf{Path-II} and \textbf{Path-III}. Under Path-II, we extract cross-shot evidence from CLIP-based semantic similarity~\citep{radford2021learning,cherti2023reproducible}, both globally and within/between event-coherence groups. Within-label similarity captures whether shots belonging to the same event line remain semantically coherent, while between-label similarity supports expressive and intellectual montage types that intentionally relate distinct semantic lines. These expert cues are passed to a VLM, which selects one montage subtype from the predefined taxonomy. The Path-II score is binary:
\[
C1_{\mathrm{B}} \in \{0,1\}.
\]
Under Path-III, we additionally ask a montage-recognition multiple-choice question derived from the structured prompt and score it as
\[
C1_{\mathrm{C}} \in \{0,1\}.
\]
The final score is the average of the two:
\[
C1=\frac{C1_{\mathrm{B}}+C1_{\mathrm{C}}}{2}.
\]

\paragraph{D. Transition execution.}
The D group evaluates shot-to-shot transition realization at the cinematographic, optical, and audio-video levels.

\textbf{D1 (cinematographic transition execution)} combines \textbf{Path-I}, \textbf{Path-II}, and \textbf{Path-III}. At the transition level, we first route each annotated transition type to the appropriate scoring method. For relatively structured transition types, Path-I is sufficient. Specifically, \emph{graphic match} is computed from the cosine similarity between DINOv2~\cite{oquab2023dinov2} features of the final frame before the cut and the first frame after the cut.
\emph{Match on action} uses RAFT-based~\cite{teed2020raft} motion vectors around the cut and scores their cosine similarity. \emph{Sound match} uses PANNs~\citep{kong2020panns} embeddings from the audio segments before and after the cut and again computes cosine similarity. \emph{Contrast cut} is scored as
\[
s_{\mathrm{contrast}}=1-\cos,
\]
where $\cos$ is the DINOv2 cross-cut frame similarity.

For more semantic transition types, we use \textbf{Path-II}. Examples include \emph{logical cut}, \emph{POV cut}, \emph{exit-and-entry transition}, \emph{wipe-by transition}, \emph{empty-shot transition}, \emph{whip-pan transition}, and \emph{extreme scale cut}. These rely on expert cues from multiple specialized models. For instance, \emph{whip-pan transition} uses RAFT optical-flow magnitude evidence; \emph{extreme scale cut} uses a MovieShots-style shot-scale estimator~\citep{rao2020unified}; \emph{POV cut} uses head-pose evidence from \textbf{6DRepNet}~\citep{hempel20226d} together with object and person detections from \textbf{YOLO}~\citep{redmon2016you}; \emph{wipe-by} uses saliency and occlusion masks~\citep{hou2007saliency}; and \emph{empty-shot transition} relies on object-detection and scene-category cues. These expert signals are then interpreted by a VLM to produce a binary judgment for that transition.

Finally, under \textbf{Path-III}, we ask a transition-type recognition question. Let $\bar{D1}_{\mathrm{trans}}$ denote the mean transition-level score from Path-I and Path-II over all transitions, and let $D1_{\mathrm{qa}}$ denote the question-answering score from Path-III. The final D1 score is
\[
D1=\frac{\bar{D1}_{\mathrm{trans}} + D1_{\mathrm{qa}}}{2}.
\]

\textbf{D2 (optical transition type)} uses \textbf{Path-I}. It is computed directly from the transition classifier associated with TransNetV2~\citep{soucek2024transnet}, which predicts one of five normalized classes: \texttt{hard\_cut}, \texttt{dissolve}, \texttt{wipe}, \texttt{flash\_white}, and \texttt{flash\_black}. 
It should be noted that TransNetV2 can only distinguish between hard cuts and dissolves, estimating confidence based on the overall intensity of visual change; continuous camera motion or intra-shot exposure change do not trigger it. Then, we further optimize our method based on this capability.
If TransNetV2 identifies a transition as a hard cut, it is labeled as a hard cut directly. If TransNetV2 predicts a dissolve, we first examine whether the entire frame changes uniformly over time. If the change is not uniform, the transition is classified as a wipe, because wipe transitions typically preserve clear regions from both the preceding and the following shots simultaneously, resulting in spatially non-uniform change across the frame. If the frame does change uniformly, we further inspect the lighting effect of the image to determine whether a full-white or full-black flash occurs; if so, the transition is classified as flash-to-white or flash-to-black, respectively. Only when none of these conditions is satisfied do we classify the transition as a dissolve.

A correct match receives $1.0$, confusion between \texttt{dissolve} and \texttt{wipe} receives partial credit $0.5$, and all other mismatches receive $0$. The final score is the mean over all evaluable transitions:
\[
D2=\frac{1}{T}\sum_{t=1}^{T} s_t.
\]
If the transition is not generated at all, it is assigned a score of $0$ directly.

\textbf{D3 (audio-video transition relation)} uses \textbf{Path-II}, although it is supported by extensive expert-model evidence. The target is whether the boundary realizes the annotated \emph{J-cut}, \emph{L-cut}, or \emph{straight cut}. We adopt an \emph{actual-video-first} strategy: we first infer the actual relation from the generated clip and then compare it against the annotation.

The signal pipeline combines \textbf{Demucs}~\citep{rouard2023hybrid} source separation, \textbf{Whisper}~\citep{radford2023robust} ASR on the vocals track, \texttt{librosa}~\cite{mcfee2015librosa} acoustic descriptors, and \textbf{PANNs}~\citep{kong2020panns} object-sound tagging. Around each cut, we analyze three sources of evidence: ambient continuity, speech continuity/projection, and salient object sounds. A signal-gating module determines whether the boundary exhibits synchronized change, candidate overlap, ambient continuity only, or unclear evidence. Before sending the evidence to the VLM, ambiguous overlap cases are normalized toward \emph{straight} so that weak or diffuse cross-fades are not over-interpreted as J/L-cuts.

A further signal-based arbitration layer is then applied without looking at the ground truth. This layer takes precedence over the VLM when the signal evidence is decisive, for example when there is a clear sync change with no substantial object-sound tail, or when an object sound clearly extends across the cut and supports either J- or L-cut. If no decisive arbitration is available, we use the VLM prediction whenever it is not \texttt{unclear}.
Let $\hat{r}_t$ be the final predicted relation and $r_t^{\mathrm{gt}}$ the normalized ground-truth relation for transition $t$. Then
\[
s_t=
\begin{cases}
1, & \hat{r}_t = r_t^{\mathrm{gt}},\\
0, & \text{otherwise}.
\end{cases}
\qquad
D3=\frac{1}{T}\sum_{t=1}^{T} s_t.
\]

\paragraph{E. Cinematography and physical plausibility.}
The E group evaluates shot-level visual execution quality.

\textbf{E1 (camera-parameter execution)} combines \textbf{Path-II} and \textbf{Path-III}. It is evaluated per aligned shot along four sub-dimensions: camera motion, shot scale, camera angle, and depth-related property. For \emph{camera motion}, we use \textbf{MonST3R}~\citep{zhang2025monst3r} with the \textbf{DUSt3R} geometry backbone~\citep{wang2024dust3r} to estimate 6-DoF camera trajectory and focal-length change, and then pass the geometric evidence to a VLM. Exact matches score $1.0$, near matches score $0.5$, and mismatches score $0$. For \emph{shot scale}, we use MovieShots-style scale evidence~\citep{rao2020unified} together with person/object detections from YOLO~\citep{redmon2016you}. If the predicted and ground-truth scales differ by $0$, $1$, $2$, or at least $3$ levels, the score is $1.0$, $0.7$, $0.3$, or $0$, respectively. For \emph{angle}, we use head-pose cues from 6DRepNet~\citep{hempel20226d} and VLM selection over the predefined angle taxonomy; this is scored as exact-match binary accuracy. For \emph{depth of field}, we use saliency and blur-based cues, including Laplacian-variance sharpness measures~\citep{pech2000diatom}, and map the result through a VLM with exact-match or same-group partial credit.

Let $E1_{\mathrm{motion}}$, $E1_{\mathrm{scale}}$, $E1_{\mathrm{angle}}$, and $E1_{\mathrm{dof}}$ denote the four mean sub-scores over aligned shots. The Path-II score is
\[
E1_{\mathrm{B}}=
\frac{
E1_{\mathrm{motion}}+
E1_{\mathrm{scale}}+
E1_{\mathrm{angle}}+
E1_{\mathrm{dof}}
}{4}.
\]
Under \textbf{Path-III}, we additionally ask a camera-parameter-combination question, yielding $E1_{\mathrm{C}}$. The final score is
\[
E1=\frac{E1_{\mathrm{B}} + E1_{\mathrm{C}}}{2}.
\]

\textbf{E2 (image quality)} uses \textbf{Path-I}. We adopt a frame-level aesthetic predictor based on \textbf{OpenCLIP ViT-L/14} with a \textbf{LAION} aesthetic linear head~\citep{schuhmann2022laion}, following the implementation style of VBench~\citep{huang2024vbench}. The model outputs an aesthetic-quality score for each frame, and the final metric is their average clipped to $[0,1]$:
\[
E2=\mathrm{clip}_{[0,1]}(\mathrm{aesthetic\_quality}).
\]

\textbf{E3 (cross-shot physical consistency)} uses \textbf{Path-III} only. It is evaluated as a VLM-based question answering task focused specifically on cross-shot physical plausibility. The question asks whether a selected physical consistency phenomenon spanning at least two shots is physically valid in the generated video. The output is binary, and we use
\[
E3 \in \{0,1\}.
\]
This question covers a broad range of physical consistency judgments, including physical errors (e.g., violations of mechanics or optics) and abrupt changes in object attributes (e.g., color or shape). 
The main error types observed in our samples include shape changes, object interpenetration, and implausible motion, all of which are evaluated under this dimension. We emphasize that our goal is not to exhaustively assess every fine-grained aspect of physical consistency, as these have already been extensively studied in prior benchmarks \citep{xie2025phyavbench,cui2026joint}. Instead, we introduce this dimension to ensure that physical quality is not overlooked when evaluating editing ability.

\paragraph{F. Audio realization.}
The F group evaluates within-shot synchronization and overall audio quality.

\textbf{F1 (within-shot audio--visual synchronization)} uses \textbf{Path-I}. We derive motion-energy peaks from RAFT~\citep{teed2020raft} optical flow and audio onset peaks from \texttt{librosa}~\citep{mcfee2015librosa}. For each motion-energy peak, we find the nearest audio onset and compute the temporal offset. 
This linear calibration is intentionally less harsh than an exponential penalty, because large-offset tails otherwise suppress scores too aggressively. 

\textbf{F2 (overall audio quality)} uses \textbf{Path-I}. We employ an audio signal-analysis service based on \texttt{librosa}~\citep{mcfee2015librosa} features. It computes an objective quality score from signal-to-noise ratio, dynamic range, spectral richness, clipping penalty, and silence penalty. The final score is simply
\[
F2=\mathrm{clip}_{[0,1]}(\mathrm{signal\_score}).
\]

\paragraph{Final aggregation.}
After all dimensions are computed, the final benchmark report keeps the 16 dimension scores separately and also reports an overall average. The combination rules follow the pathway design described above: A1, A4, B1, B2, D2, E2, F1, and F2 come directly from \textbf{Path-I}; B3, C1, D1, D3, and E1 come from \textbf{Path-II}; A2, A3, E3, and the question-answering components of C1, D1, and E1 come from \textbf{Path-III}.

The benchmark-wide overall score is the arithmetic mean of the 16 final dimension scores. This design ensures that each metric is computed using the type of evidence most appropriate to the editing phenomenon it measures.

\subsection{Metrics Validation}

\subsubsection{Shots Pre-process Ablation and Human Alignment}
\label{app:shot_alignment_ablation}

Since some of our evaluation dimensions are highly correlated with the shot alignment results obtained during shot preprocessing, we conduct a correlation analysis between the accuracy of VLM shot alignment and human judgment.

\begin{wraptable}{r}{0.35\textwidth}
    \vspace{-1.5em}
    \centering
    \caption{Shot alignment precision, recall and F1.}
    \resizebox{0.33\textwidth}{!}{
    \begin{tabular}{c|c|c}
    \toprule
       Precision  & Recall & F1 Score \\
    \midrule
       92.54\%  & 95.48\% & 93.99\% \\
    \bottomrule
    \end{tabular}
    }
    \label{app:tab_statis_shot_align}
\end{wraptable}

\begin{figure}[h!]
    \centering
    \includegraphics[width=1\linewidth]{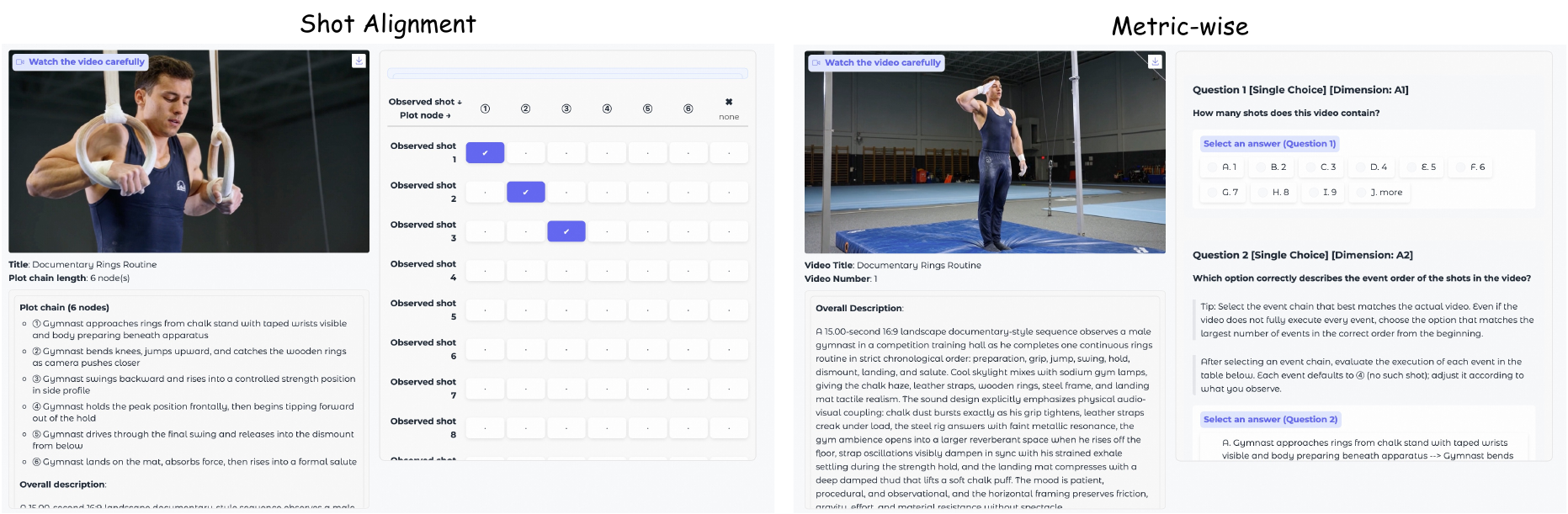}
    \caption{Screenshot of shot alignment and metric-wise alignment questionnaire}
    \label{app:fig_human3}
\end{figure}

We invited human experts to directly assess the relationship between shots and plot in the videos. The questionnaire asked human annotators to choose the matching relationship between shots and plot based on the video content. A screenshot of the questionnaire is shown in Figure \ref{app:fig_human3}. We report the precision, recall, and F1 scores of automated shot alignment against human evaluation in Table \ref{app:tab_statis_shot_align}.
The results prove that our shot alignment procedure is highly consistent with human judgment.

In addition, we further examine the impact of incorporating shot alignment on the final results. We conduct ablation study on the human-annotated dataset.
The results are reported in Table \ref{app:tab_human_align_shot_align}.

\begin{table}[h!]
    \centering
    \caption{Ablations on with or without shot alignment.}
    \resizebox{0.6\textwidth}{!}{%
    \begin{tabular}{c|c|ccccc}
    \toprule
        Correlation&  Ablation & A2 & A3 & C1 & D1 & D3 \\
    \midrule
        \multirow{2}*{Pearson $\uparrow$}  & w/ shot align  & 0.94 & 0.93 & 0.85 & 0.92 & 0.97  \\
        ~  & w/o shot align  & 0.93 & 0.84 & 0.74 & 0.45 & 0.87  \\
        \midrule
        
        \multirow{2}*{Spearman $\uparrow$} & w/ shot align& 1.00  & 0.80 & 0.80 & 0.95 & 0.95  \\
        ~ & w/o shot align & 1.00 & 0.20 & 0.80 & 0.40 & 0.20  \\

    \bottomrule
    \end{tabular}%
    }
    \label{app:tab_human_align_shot_align}
\end{table}

The results demonstrate the necessity of shot alignment. Without this processing step, fragmented shots generated by the model or missing shots that cause misalignment in the overall event chain can adversely affect subsequent evaluation results. This is especially true for metrics related to event chains, montage, and transitions, as these metrics require accurate shot localization and then use the prompt instructions at the corresponding positions for evaluation. If misalignment occurs, it leads to artificially low scores.

\subsubsection{Metric-wise Human Alignment}
\label{app:human_aligment}

To verify the alignment between several of our editing-centric metrics and human perception, we conduct a detailed human-alignment study. 
All human annotators in this study have relevant professional backgrounds in film editing or related audiovisual production. Most have received formal training in film, television, digital media, or closely related fields, and all have at least 3 years of practical experience in video editing, post-production, or cinematographic work.
Specifically, we evaluate the dimensions A1, A2, A3, C1, D1, D3, and E3.

Considering that the other dimensions have already been covered by existing benchmarks or have been widely used and validated, and given the high human labor cost of metric-wise annotation, we focus our evaluation only on the core editing-related metrics.
Human experts thoroughly studied the prior knowledge consistent with the instructions provided to the VLM before the evaluation, so as to ensure that the response process was minimally affected by personal preferences. During the questionnaire process, the order of samples was randomly shuffled to prevent dependencies caused by a fixed order.


The \textbf{A1} dimension measures the number of shots, requiring annotators to count the visually perceived shots in the video. All visually observable shot changes should be counted, including abrupt jumps caused by frame dropping. This serves as an important indicator of the continuity and completeness of shot generation.
The \textbf{A2} dimension evaluates the event chain. Consistent with the VLM-based protocol, human annotators are first asked to select the event chain that best matches the video, and then to judge the degree of execution for each step in that chain.
The \textbf{A3} dimension requires human annotators to determine the editing logic for each transition by selecting the most appropriate option from four choices. The \textbf{C1} dimension requires identifying the type of montage. The \textbf{D1} and \textbf{D3} dimensions require selecting the camera dimension and the audio-video relationship for each transition. The \textbf{E3} dimension requires judging whether the generated video satisfies consistency requirements, such as adherence to physical laws, during the editing process.
We aggregate the final human evaluation results using the same scoring scheme as CutCraft, and compare the resulting scores on the relevant dimensions with those from the automated evaluation. 
The results are already shown in Table \ref{tab:human_align}. 
Additionally, the Pearson and Spearman correlation coefficients for the A1 dimension which is unrelated to VLM, are 0.91 and 0.95.

In addition, to demonstrate the consistency among our human experts, we performed a statistical analysis of the agreement in their questionnaire responses. For each question on each sample, we defined the agreement ratio as the proportion of annotators choosing the majority answer among all responses. We then computed the average agreement across all evaluated dimensions. The results are reported in Table \ref{app:tab_human_consi}.
As can be seen, the agreement among human annotators is above 70\% for all dimensions. The more challenging dimensions, including C1 montage type and D1 cinematographic transition type, show relatively lower agreement, but still reach a clear majority consensus and are therefore unlikely to substantially affect the final results.

\begin{table}[h!]
    \centering
    \caption{Consistency analysis of human experts.}
    \resizebox{0.65\textwidth}{!}{%
    \begin{tabular}{c|ccccccc}
    \toprule
        - &  A1 & A2 & A3 & C1 & D1 & D3 & E3 \\
    \midrule
        \multirow{1}*{Consistency $\uparrow$}  & 94.05\% & 88.44\% & 81.85\% & 77.84\% & 72.48\% & 83.68\% & 84.92\% \\      
 
    \bottomrule
    \end{tabular}%
    }
    \label{app:tab_human_consi}
\end{table}

\subsubsection{Testing with Positive and Negative Samples}
\label{app:posi_neg}

\paragraph{Sample Displacement Test.}
We first conducted a sample displacement test. Specifically, we mismatched the videos generated by the high-performing Seedance 2.0 model with their corresponding prompts by shifting them by 30 samples, and then fed the misaligned pairs back into the evaluation framework to analyze the resulting changes in evaluation scores.
The result is presented in Table \ref{app:tab_displacement}.

\begin{table*}[ht!]
\centering
\caption{Displacement test on Seedance 2.0.}
\label{app:tab_displacement}
\resizebox{\textwidth}{!}{%
\begin{tabular}{l|cccc|ccc|c|ccc|ccc|cc|c}
\toprule

\multirow{2}*{\textbf{Model}} &  \multicolumn{4}{c|}{\textbf{Narrative}} & \multicolumn{3}{c|}{\textbf{Rhythm \& pacing}} & \multicolumn{1}{c|}{\textbf{Montage}} & \multicolumn{3}{c|}{\textbf{Transition}} &
\multicolumn{3}{c|}{\textbf{Cinematography}} & \multicolumn{2}{c|}{\textbf{Audio}} & \multirow{2}*{\textbf{Overall $\uparrow$}} \\

~ & A1$\uparrow$ & A2$\uparrow$ & A3$\uparrow$ & A4$\uparrow$ & B1$\uparrow$ & B2$\uparrow$ & B3$\uparrow$ & C1$\uparrow$ & D1$\uparrow$ & D2$\uparrow$ & D3$\uparrow$ & E1$\uparrow$ & E2$\uparrow$ & E3$\uparrow$ & F1$\uparrow$ & F2$\uparrow$ & ~ \\
\midrule

Origin & 0.865 & 0.800 & 0.712 & 0.653 & 0.739 & 0.571 & 0.653 & 0.490 & 0.440 & 0.584 & 0.382 & 0.561 & 0.539 & 0.829 & 0.687 & 0.555 & 0.629 \\

Displacement & 0.605 & 0.003 & 0.003 & 0.126 & 0.003 & 0.432 & 0.617 & 0.164 & 0.045 & 0.003 & 0.003 & 0.033 & 0.538 & 0.089 & 0.729 & 0.565 & 0.247\\

\bottomrule
\end{tabular}%
}
\end{table*}

The results show that the dimensions highly correlated with camera-shot prompts and transition localization (e.g., \textbf{A1} shot-count, \textbf{A2} event execution, \textbf{A3} causal coherence \textbf{A4} style, \textbf{B1} shot duration, \textbf{C1} montage, \textbf{D1}  cinematographic transition, \textbf{D2} optical transition , \textbf{D3} transition audio-video relation, and \textbf{E3} physical consistency) exhibit substantial score drops after misalignment.
In contrast, only prompt-independent general dimensions, such as \textbf{B3} (rhythm–mood matching), \textbf{E2} (image quality), \textbf{F1} (audio–video synchronization) and \textbf{F2} (audio quality) remain largely unchanged. These findings further demonstrate the soundness of our evaluation metric design.

\paragraph{Manual Positive and Negative Sample Test of Transition Related Metrics.}
We present qualitative results for some metrics by manually constructing positive and negative samples. In common video editing software, the most typical manual operations include transitions between shots, the handling of audio and video tracks, and the concatenation of different numbers of shots. Since the number of shots has already been thoroughly evaluated in the preceding experiments, we do not construct positive and negative samples for it here.

\begin{figure}[h!]
    \centering
    \includegraphics[width=0.8\linewidth]{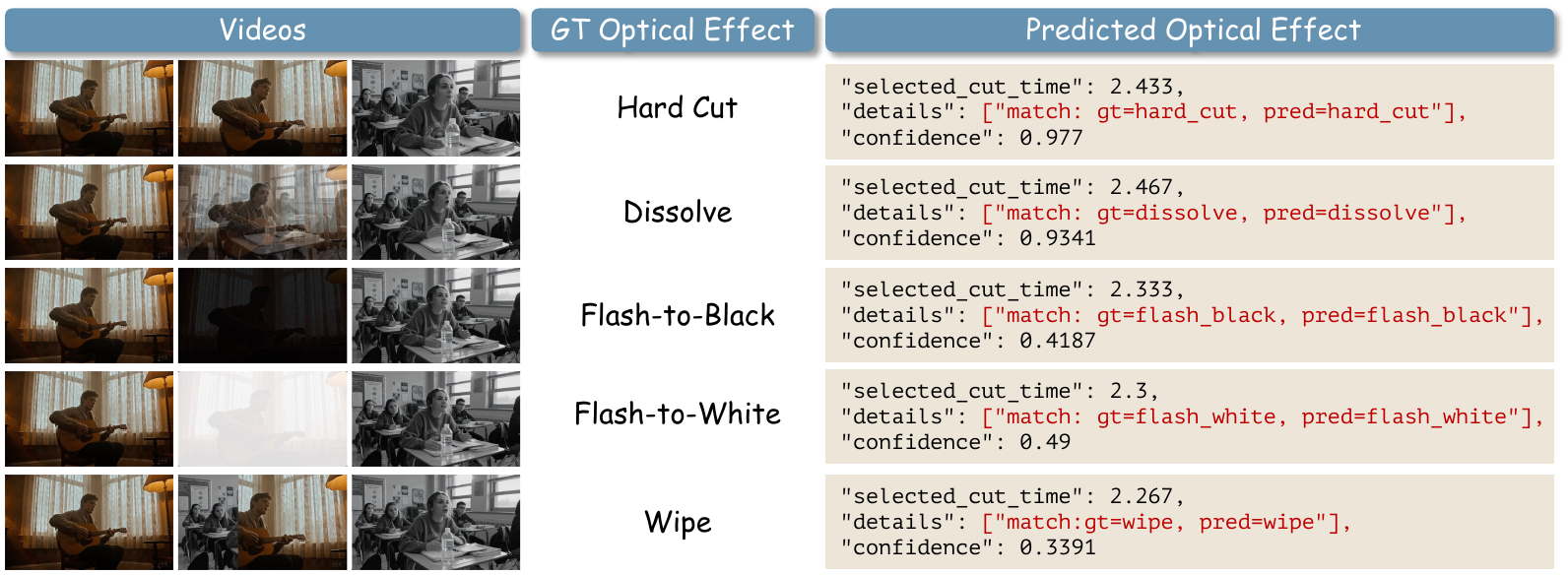}
    \caption{Positive and negative sample tests for manually specified transition optical effects.
}
    \label{app:fig_posi_nega_optical}
\end{figure}

We first evaluated the \textbf{D2} optical transition type dimension. We concatenated two video clips and then manually applied transition effects at the junction, including hard cuts, dissolves, flash-to-black, flash-to-white, and wipes. We then ran the detection for the \textbf{D2} dimension alone. The results are shown in Figure \ref{app:fig_posi_nega_optical}. As can be seen, our method is able to accurately identify different types of transition effects and can determine the transition timing with near-precise accuracy based on the intensity of content changes between shots.

\begin{figure}[h!]
    \centering
    \includegraphics[width=0.8\linewidth]{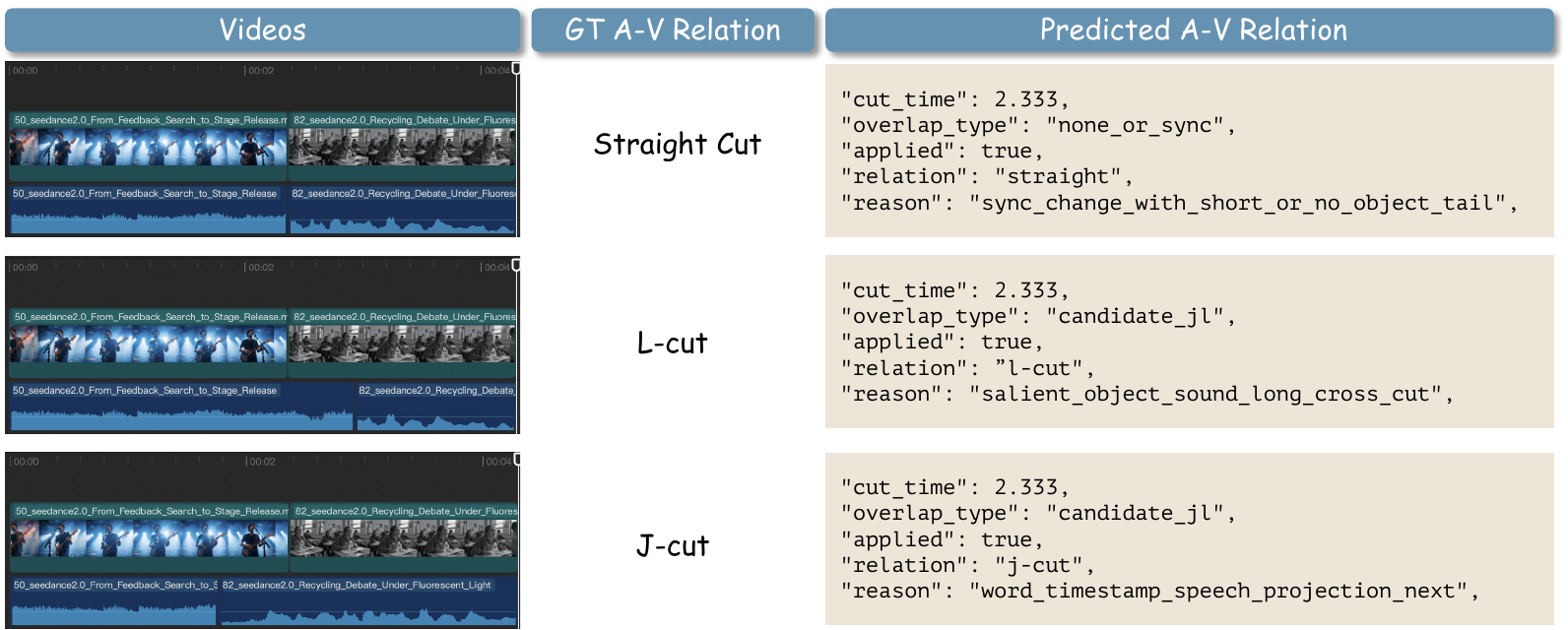}
    \caption{Positive and negative sample tests for manually specified transition audio-video relation.}
    \label{app:fig_posi_nega_av}
\end{figure}

We also evaluated the \textbf{D3} dimension, which focuses on the audio-video relationship across transitions. We selected two video clips and manually adjusted the overlap between their audio and video tracks in editing software, after which we conducted the \textbf{D3} evaluation independently. The results are shown in Figure \ref{app:fig_posi_nega_av}. As can be seen, our evaluation method can accurately identify changes in audio-video relationships, demonstrating the accuracy of our approach.

\subsubsection{Metric-wise Sensitivity Analysis}
\label{app:sensitivity}

We conduct a sensitivity analysis of the manually specified parameters in the evaluation metrics, focusing mainly on the event-execution score setting in A2 and the $\lambda_0$ parameter in the causal chain metric A3. The current scoring scheme for A2 consists of four levels, $[1.0, 0.5, 0.1, 0]$, noted as Group 1. For comparison, we additionally introduce two alternative schemes: $[1.0, 0.6, 0.2, 0]$ (Group 2) and $[1.0, 0.7, 0.4, 0.1]$ (Group 3). The current value of $\lambda_0$ in A3 is 0.3; we further evaluate three additional settings, namely 0.2, 0.4, and 0.5.

\begin{table}[h!]
    \centering
    \caption{Metric-wise Sensitivity Analysis.}
    \resizebox{0.9\textwidth}{!}{%
    \begin{tabular}{l|cccccc|cccccccc}
    \toprule
        \multirow{3}*{\textbf{Model}} & \multicolumn{6}{c|}{\textbf{A2 score}} & \multicolumn{8}{c}{\textbf{A3 score}} \\
        \cmidrule(lr){2-7} \cmidrule(lr){8-15}
        ~ & \multicolumn{2}{c}{Group 1} & \multicolumn{2}{c}{Group 2} & \multicolumn{2}{c|}{Group 3} & \multicolumn{2}{c}{$\lambda_0=0.2$} & \multicolumn{2}{c}{$\lambda_0=0.3$} & \multicolumn{2}{c}{$\lambda_0=0.4$} & \multicolumn{2}{c}{$\lambda_0=0.5$} \\
        \cmidrule(lr){2-3} \cmidrule(lr){4-5} \cmidrule(lr){6-7} \cmidrule(lr){8-9} \cmidrule(lr){10-11} \cmidrule(lr){12-13} \cmidrule(lr){14-15}
        ~ & Score & Rank & Score & Rank & Score & Rank & Score & Rank & Score & Rank & Score & Rank & Score & Rank \\
    \midrule
        Minimax H3     & 0.815 & 1  & 0.845 & 1  & 0.883 & 1  & 0.703 & 1  & 0.729 & 1  & 0.755 & 1  & 0.781 & 1 \\
        Seedance2.5    & 0.766 & 3  & 0.798 & 3  & 0.843 & 3  & 0.680 & 4  & 0.705 & 4  & 0.730 & 4  & 0.755 & 4 \\
        Seedance2.0    & 0.800 & 2  & 0.829 & 2  & 0.868 & 2  & 0.686 & 3  & 0.712 & 3  & 0.737 & 3  & 0.762 & 3 \\
        Happyhorse1.1  & 0.643 & 5  & 0.675 & 5  & 0.726 & 5  & 0.663 & 5  & 0.689 & 5  & 0.715 & 5  & 0.741 & 5 \\
        Kling V3       & 0.737 & 4  & 0.774 & 4  & 0.823 & 4  & 0.690 & 2  & 0.716 & 2  & 0.742 & 2  & 0.769 & 2 \\
        Veo3.1         & 0.441 & 7  & 0.470 & 7  & 0.531 & 7  & 0.490 & 7  & 0.512 & 7  & 0.535 & 7  & 0.558 & 7 \\
        Wan2.7         & 0.451 & 6  & 0.480 & 6  & 0.537 & 6  & 0.524 & 6  & 0.549 & 6  & 0.574 & 6  & 0.599 & 6 \\
        Vidu Q3        & 0.398 & 8  & 0.423 & 8  & 0.482 & 8  & 0.409 & 8  & 0.432 & 8  & 0.456 & 8  & 0.479 & 8 \\
        LTX2.3         & 0.291 & 9  & 0.315 & 9  & 0.377 & 9  & 0.369 & 9  & 0.390 & 9  & 0.412 & 9  & 0.433 & 9 \\
        MOVA           & 0.151 & 10 & 0.182 & 10 & 0.258 & 10 & 0.278 & 10 & 0.296 & 10 & 0.313 & 10 & 0.331 & 10 \\
        Ovi            & 0.078 & 12 & 0.093 & 12 & 0.158 & 11 & 0.077 & 11 & 0.081 & 11 & 0.085 & 11 & 0.089 & 11 \\
        Davinci        & 0.094 & 11 & 0.104 & 11 & 0.157 & 12 & 0.027 & 13 & 0.029 & 13 & 0.031 & 13 & 0.033 & 13 \\
        JavisDiT++     & 0.059 & 13 & 0.073 & 13 & 0.138 & 13 & 0.056 & 12 & 0.062 & 12 & 0.068 & 12 & 0.074 & 12 \\
    \bottomrule
    \end{tabular}%
    }
    \label{app:tab_sensitivity}
\end{table}

Table~\ref{app:tab_sensitivity} presents the sensitivity analysis of the manually specified parameters in A2 and A3. Overall, although different parameter settings lead to changes in the absolute scores, they have little effect on the relative ranking of models within each dimension. For A2, increasing the scores assigned to intermediate execution levels consistently raises the overall scores of all models, yet the ranking remains largely unchanged. A similar trend can be observed for A3: as $\lambda_0$ increases from 0.2 to 0.5, the scores of all models increase accordingly, while the ordering among models is well preserved. These results indicate that our evaluation metrics are robust to reasonable variations in manually chosen parameters, and that the comparative conclusions drawn within each dimension are stable.

\subsubsection{Overall Score Aggregation Strategy}
\label{app:aggregation}

Since our benchmark is designed not only to evaluate video models’ ability to follow professional editing instructions, but also to assess several general metrics that reflect overall video quality, the main text directly aggregates all dimensions into a single score due to space limitations. 
Here, we adopt two different aggregation strategies to analyze the capabilities of different models at different levels.
The first aggregation strategy groups the metrics directly according to the top-level categories in our dimension design (\textbf{A.} Narrative execution, \textbf{B.} Rhythm and pacing, \textbf{C.} Montage execution, \textbf{D.} Transition execution, \textbf{E.} Cinematography and physical plausibility, \textbf{F.} Audio realization), denoted as ``Six Groups.'' The second aggregation strategy divides the metrics into editing-intensive dimensions (\textbf{A1} shot-count, \textbf{A2} event execution, \textbf{A3} causal-chain, \textbf{B1} shot-duration, \textbf{C1} montage, \textbf{D1} cinematographic transition execution, \textbf{D2} optical transition type, \textbf{D3} audio–video transition relation) and other supporting dimensions, denoted as ``Two Groups.''
The results are presented in Table \ref{app:tab_aggregation}.

\begin{table}[h!]
    \centering
    \caption{Different Aggregation Strategy of Scores.}
    \resizebox{0.65\textwidth}{!}{%
    \begin{tabular}{l|cccccc|cc}
    \toprule
        \multirow{2}*{\textbf{Model}} & \multicolumn{6}{c|}{\textbf{Six Groups}} & \multicolumn{2}{c}{\textbf{Two Groups}} \\
        & A & B & C & D & E & F & Editability & Supporting \\
    \midrule
    Minimax H3 & 0.782 & 0.726 & 0.502 & 0.421 & 0.666 & 0.565 & 0.649 & 0.625 \\
    Seedance2.5 & 0.754 & 0.705 & 0.498 & 0.423 & 0.644 & 0.642 & 0.621 & 0.643 \\
    Seedance2.0 & 0.757 & 0.654 & 0.490 & 0.469 & 0.643 & 0.621 & 0.626 & 0.631 \\
    Happyhorse1.1 & 0.705 & 0.684 & 0.458 & 0.416 & 0.640 & 0.603 & 0.589 & 0.624 \\
    Kling V3 & 0.765 & 0.689 & 0.454 & 0.395 & 0.638 & 0.519 & 0.620 & 0.595 \\
    Veo3.1 & 0.587 & 0.564 & 0.371 & 0.275 & 0.584 & 0.669 & 0.436 & 0.605 \\
    Wan2.7 & 0.617 & 0.517 & 0.431 & 0.297 & 0.558 & 0.474 & 0.460 & 0.535 \\
    Vidu Q3 & 0.518 & 0.464 & 0.391 & 0.220 & 0.501 & 0.637 & 0.353 & 0.559 \\
    LTX2.3 & 0.488 & 0.439 & 0.334 & 0.192 & 0.531 & 0.591 & 0.308 & 0.561 \\
    MOVA & 0.392 & 0.473 & 0.263 & 0.173 & 0.464 & 0.676 & 0.286 & 0.529 \\
    Ovi & 0.284 & 0.304 & 0.222 & 0.033 & 0.414 & 0.689 & 0.132 & 0.492 \\
    Davinci & 0.256 & 0.319 & 0.188 & 0.007 & 0.423 & 0.632 & 0.084 & 0.506 \\
    JavisDiT++ & 0.262 & 0.422 & 0.229 & 0.041 & 0.338 & 0.446 & 0.172 & 0.399 \\
    \bottomrule
    \end{tabular}%
    }
    \label{app:tab_aggregation}
\end{table}

Under the ``Six Groups'' aggregation, the leading models, including Minimax H3, Seedance2.5, Seedance2.0, Happyhorse1.1, and Kling V3, show consistently strong performance across most groups, especially in \textbf{A} and \textbf{E}. Among them, Minimax H3 achieves the best result in \textbf{A}, while Seedance2.0 is slightly stronger in \textbf{D}, indicating complementary strengths among top models. In contrast, \textbf{D} is the most challenging group for nearly all models, with uniformly lower scores, suggesting that transition execution remains difficult for current video generation systems. 

Under the ``Two Groups'' aggregation, the contrast between \emph{Editability} and \emph{Supporting} becomes more explicit. Top models perform well on both groups, with editability scores around 0.60 and similarly strong supporting scores. 
Specifically, for the supporting metric, Seedance 2.5 achieves the highest score, but its editability score is not the highest.
Furthermore, many mid- and lower-tier models achieve much higher scores on \emph{Supporting} than on \emph{Editability}, indicating that general video quality is easier to achieve than fine-grained editing control.

\section{Agentic Baseline Details}
\label{app:agentic_baseline}

The agentic editing baseline treats structured multi-shot prompts as editing plans. Each case is first segmented into shot-level units with target durations, then passed through global consistency analysis, shot prompt rewriting, generation planning, per-shot synthesis, and explicit post-production composition. Consistency control is driven by prompt-side labels such as \texttt{event\_coherence\_label}, which are used to group shots that share the same subject, event line, or audio reference stream. A global analysis stage infers recurring entities, settings, and storylines, and maps each shot to the corresponding identity, environment, and logical predecessor. These shared constraints are injected into rewritten shot prompts so that subject appearance, clothing, scene layout, and storyline continuity remain stable even when adjacent shots belong to different narrative threads. For shots with the same label, the first successful generation serves as a visual anchor, and later shots reuse reference frames to reduce appearance drift; when the editing structure is parallel or cross-cut, continuity is inherited from the logically previous shot on the same event line instead of the physically previous shot.

Shot generation is planned independently for each shot. The planner determines the rewritten prompt, negative prompt, generation mode, reference usage, target net duration, and the additional headroom required for editing operations. 

The composition stage executes editing annotations directly. Optical transition labels are mapped to concrete rendering operations such as hard cuts, dissolves, wipes, flash-to-white, and flash-to-black. Transition durations are normalized into executable composition parameters, and overlap-based effects consume the extra footage reserved during generation. Audio-video relations are handled on an independent audio timeline. 
The sound pipeline distinguishes clip-level diegetic audio, scene-level diegetic beds that may persist across multiple shots, and optional global background music, allowing dialogue continuity, visible sound sources, and cross-shot ambient flow to be controlled separately.

The full agent loop operates after an initial composed video is available. The rendered result is evaluated on the same editing dimensions used by the benchmark, including shot timing accuracy, transition effect execution, and audio-video relation realization. A lightweight audio gate is applied before full evaluation to inspect whether clips that need audible heads or tails for J-cuts or L-cuts are effectively silent.
Evaluation outputs are then aggregated into transition-level failure records and passed to a central planner, which attributes failures to specific transitions and produces executable repair actions. Typical actions include changing transition effect duration for visual transition failures, shifting cut timing for shot-boundary timing errors, and adjusting audio offsets or strengthening shot prompts when J-cut or L-cut realization fails. The repaired plan is then re-composed and re-evaluated, for up to two repair rounds, and the best-performing version is retained.


\section{Experimental Details}
\label{app:experiment}

\subsection{Implementation}
\label{app:implementation}

The expert models mentioned in this paper were all deployed as FastAPI-based microservices to ensure that there were no conflicts across environments. The code was deployed on a host equipped with 8$\times$A100 GPUs. The LLM used is GPT-5.4~\citep{openai2025gpt54}, and the VLM used is Qwen3.5-Omni~\citep{team2026qwen3}.
The video resolution was set to 720P; for models that do not support 720P generation, we used the closest available resolution to 720P (e.g., Minimax H3 uses 768P).

\subsection{Result Analysis}
\label{app:result}

\begin{figure}[h!]
    \centering
    \includegraphics[width=0.7\linewidth]{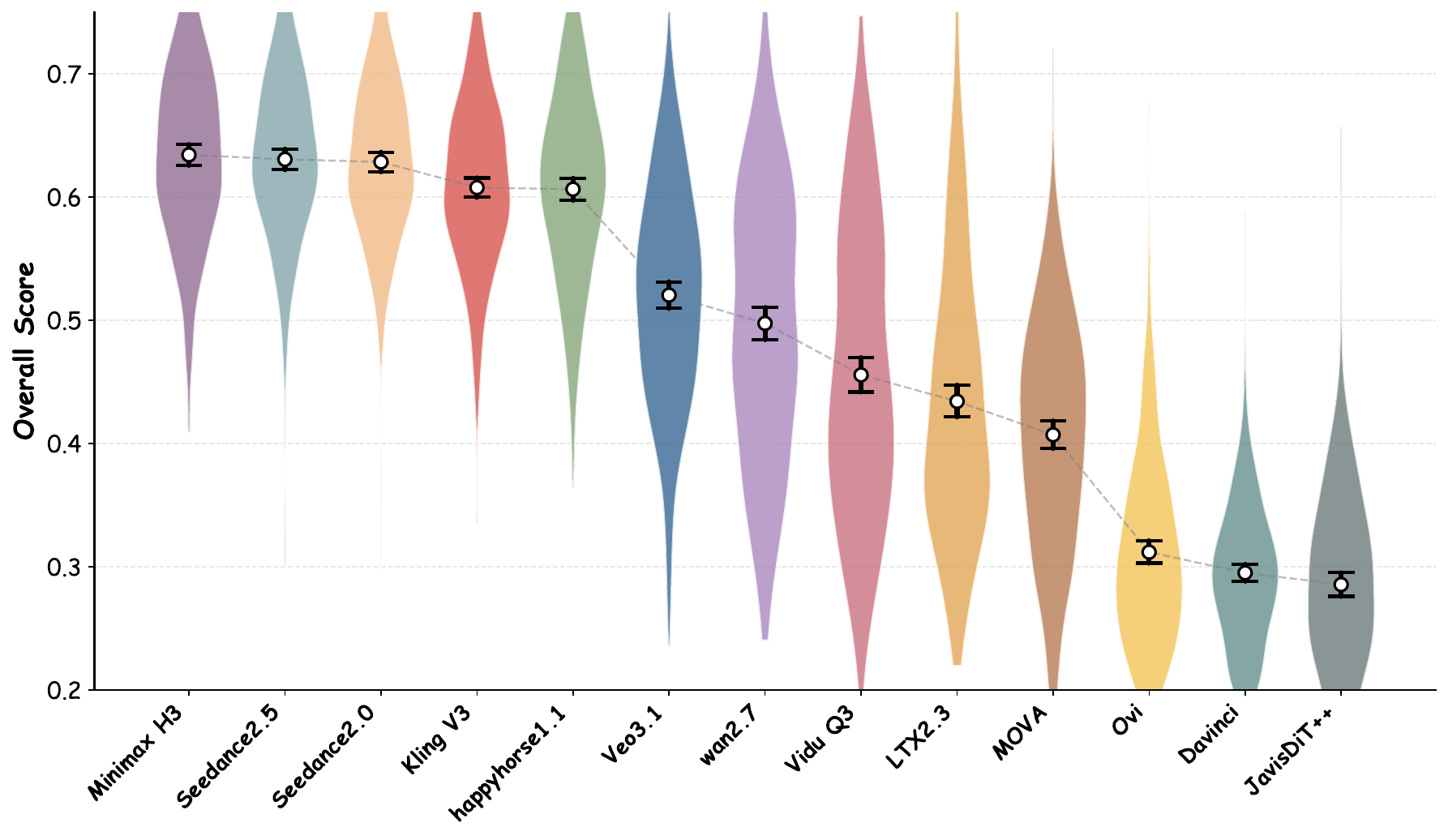}
    \caption{Overall score confidence interval analysis.}
    \label{app:fig_violin}
\end{figure}

To further assess the stability and discriminative power of the benchmark, we perform bootstrap analysis on the overall score. The results show that CutCraft can separate most models reliably: 71 out of 78 pairwise comparisons are statistically significant after Holm-Bonferroni correction, and the score distributions of models from different performance tiers are clearly separated. 
We visualize the resulting distributions of the 13 non-agent methods in Figure \ref{app:fig_violin}.
This indicates that the benchmark provides sufficient resolution for robust model-level comparison. 
Overall, the violin plot confirms that CutCraft is strongly discriminative at the benchmark level, while also reflecting a realistic phenomenon: the strongest models are easier to group into a competitive leading cluster than to rank with large statistical margins.

\subsection{More Cases}
\label{app:more_cases}

We provide additional video failure cases here, covering multiple video generation models and various types of failures.

\begin{figure}[ht!]
    \centering
    \includegraphics[width=0.7\linewidth]{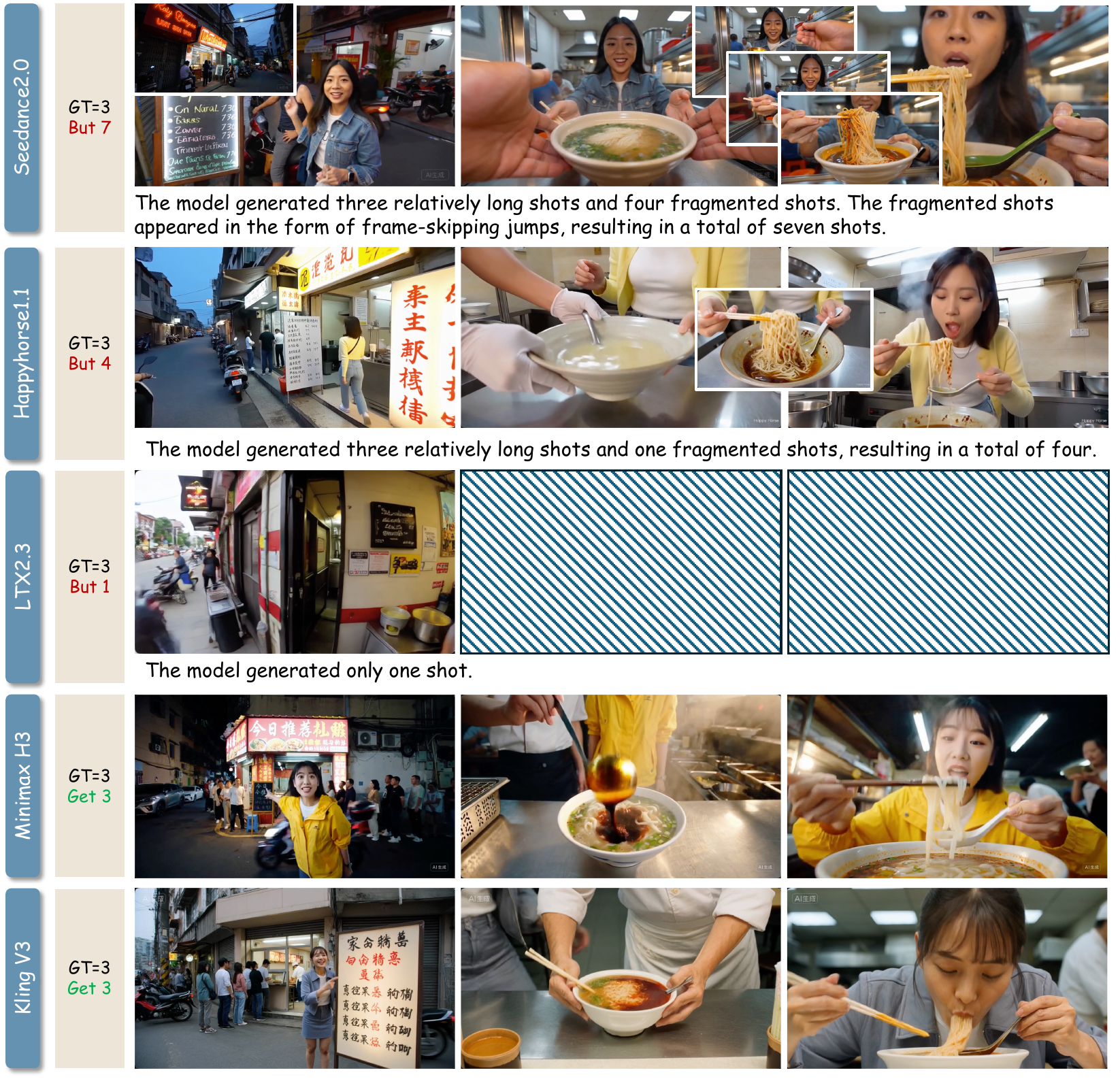}
    \caption{\textbf{Shot Count.} The GT shot count is three, but both Seedance2.0 and Happyhorse1.1 generated very short fragmented shots. In contrast, LTX2.3 produced only a single shot. Specifically, the fragmented shots in Seedance2.0 were caused by frame skipping, whereas Happyhorse1.1 directly introduced an additional viewpoint.
}
    \label{app:fig_morecase_shot1}
\end{figure}

\begin{figure}[ht!]
    \centering
    \includegraphics[width=1\linewidth]{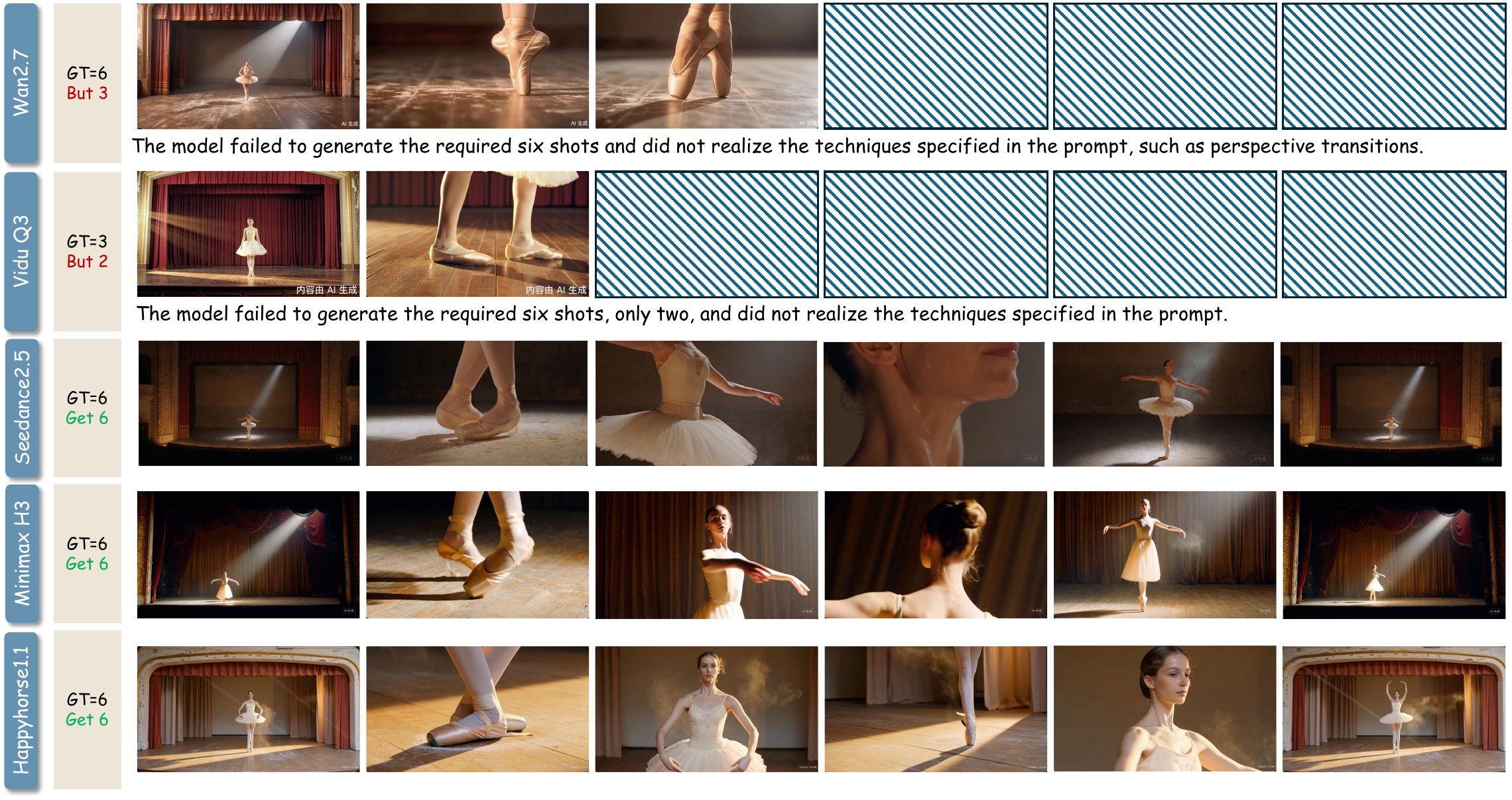}
    \caption{\textbf{Shot Count.} The GT shot count is six, but both Wan2.7 and Vidu Q3 omitted some shots. In particular, the second and third shots generated by Wan2.7 are actually two shots split from the same scene.
}
    \label{app:fig_morecase_shot2}
\end{figure}

\paragraph{A1 Shot-Count Failure.}
Here we present two groups of examples: one with a ground truth of three shots and the other with a ground truth of six shots. We observe that the failure cases mainly involve fragmented shots or the omission of shots appearing later in the sequence.
The results are presented in Figure \ref{app:fig_morecase_shot1} and \ref{app:fig_morecase_shot2}.

\begin{figure}[t!]
    \centering
    \includegraphics[width=0.77\linewidth]{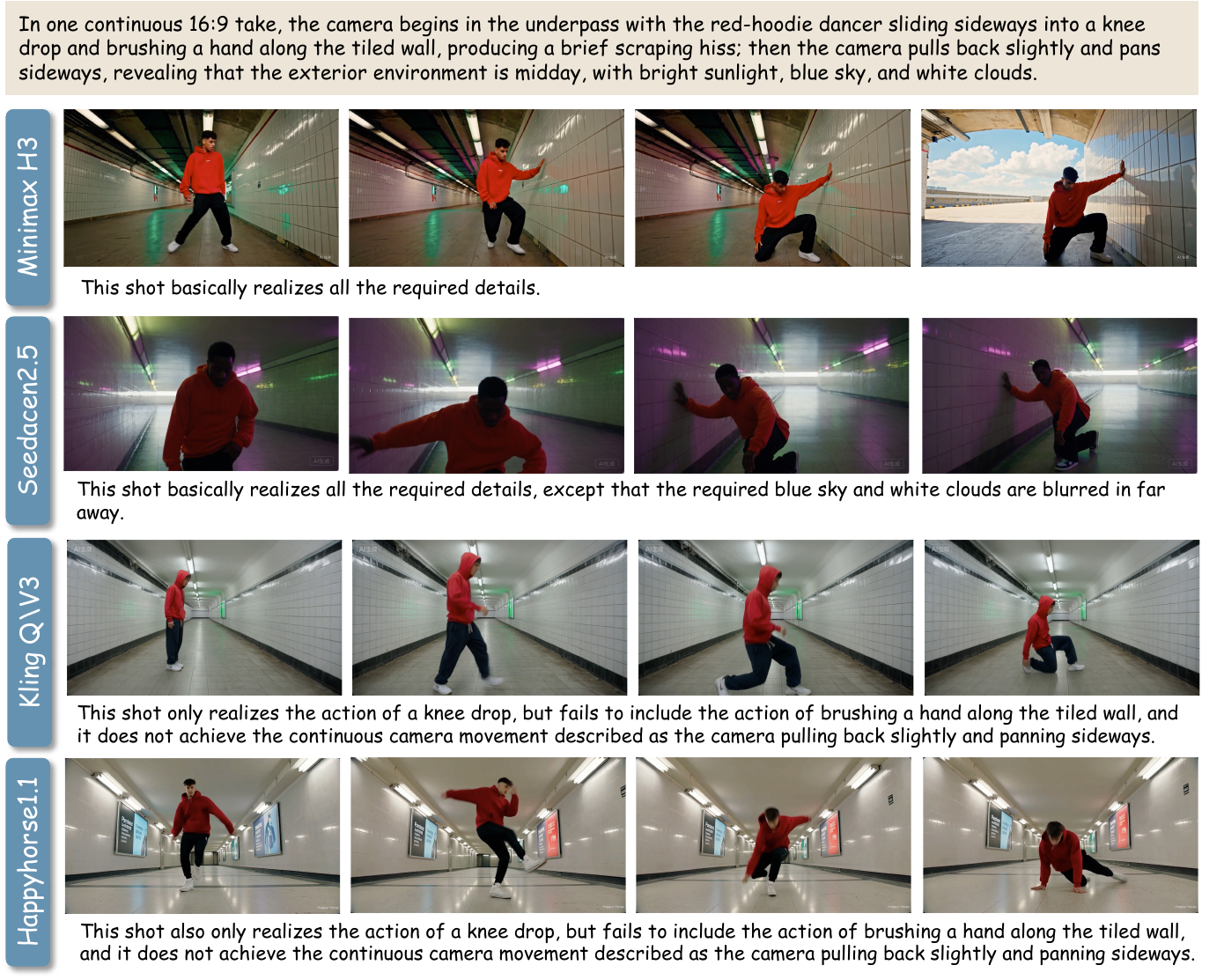}
    \caption{\textbf{Event Execution.} This example presents the content of a single shot, where the prompt describes complex dancing movements and scene-level camera motion changes. Only MiniMax H3 and Seedance2.5 perform relatively well, while the other models omit many details.
}
    \label{app:fig_morecase_event}
\end{figure}

\paragraph{A2 Event Execution Failure.}
Here we present one set of examples in which the shot description requires complex character actions and detailed scene elements, but only some models are able to accurately understand and execute the complex scene actions and camera movement instructions. The results are shown in Figure \ref{app:fig_morecase_event}.

\paragraph{A3 Causal Coherence Failure.}

\begin{figure}[t!]
    \centering
    \includegraphics[width=0.65\linewidth]{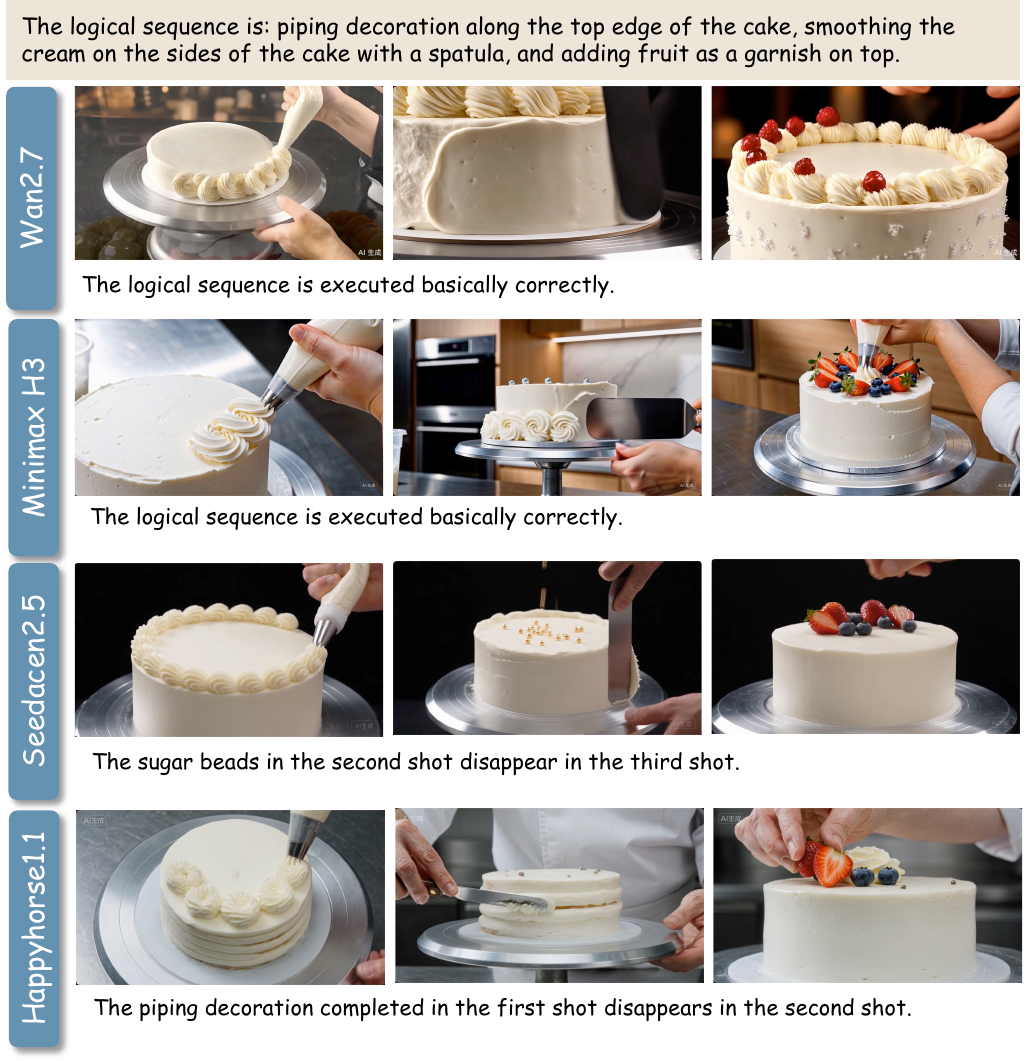}
    \caption{\textbf{Causal Coherence. }This example shows a cake-making process. Wan2.7 and Minimax H3 are generally able to realize the intended logical chain, whereas the other models break the overall logic by losing part of the details in the second shot. This is not merely a simple failure to execute an event; rather, the event execution failure affects the logic of the entire video.
}
    \label{app:fig_morecase_causal}
\end{figure}

Here we present a set of examples showing a continuous cake-making process. However, some models fail to preserve the steps completed in the previous shot when generating the next shot, resulting in errors in the overall logical chain. Note that in this dimension, we focus primarily on the logic of the events rather than on whether the frames contain physical errors or other visual failures.
The results are shown in Figure \ref{app:fig_morecase_causal}.

\paragraph{C1 Montage Type Failure.}

\begin{figure}[t!]
    \centering
    \includegraphics[width=0.7\linewidth]{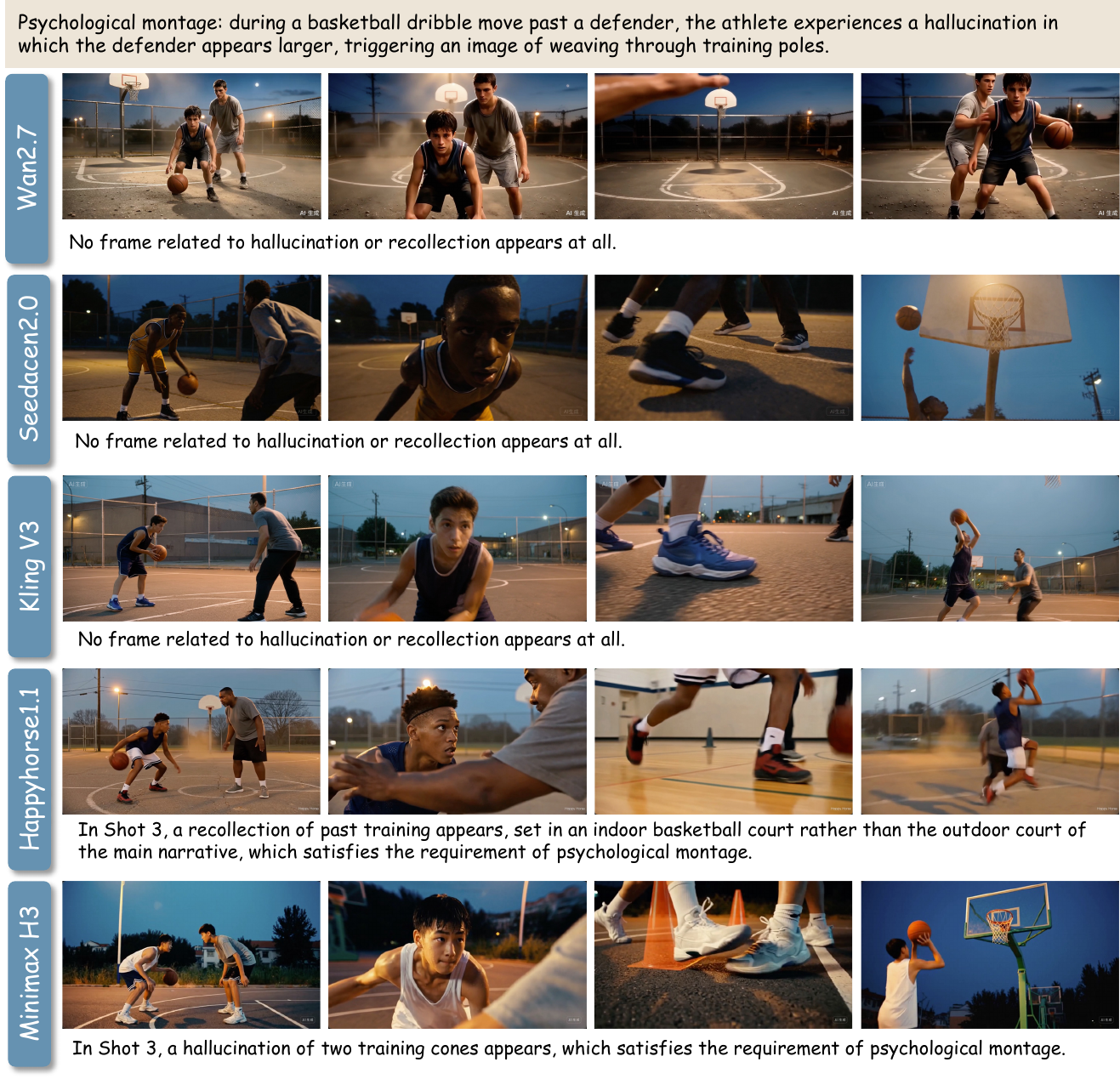}
    \caption{\textbf{Montage Type. }The prompt requires the realization of psychological montage, in which the player imagines a cone-dribbling training scene while driving past the defender in basketball. Only Happyhorse1.1 and Minimax H3 successfully realize the psychological montage.
}
    \label{app:fig_morecase_montage}
\end{figure}

Here we present a set of examples of psychological montage. For complex montage type (not sequential montage), the model needs to fully understand the editing technique and generate features that may lie outside the data distribution of the main event chain, which is challenging for the model. The results are shown in Figure \ref{app:fig_morecase_montage}.

\paragraph{D1 Cinematographic Transition Failure.}

\begin{figure}[t!]
    \centering
    \includegraphics[width=0.8\linewidth]{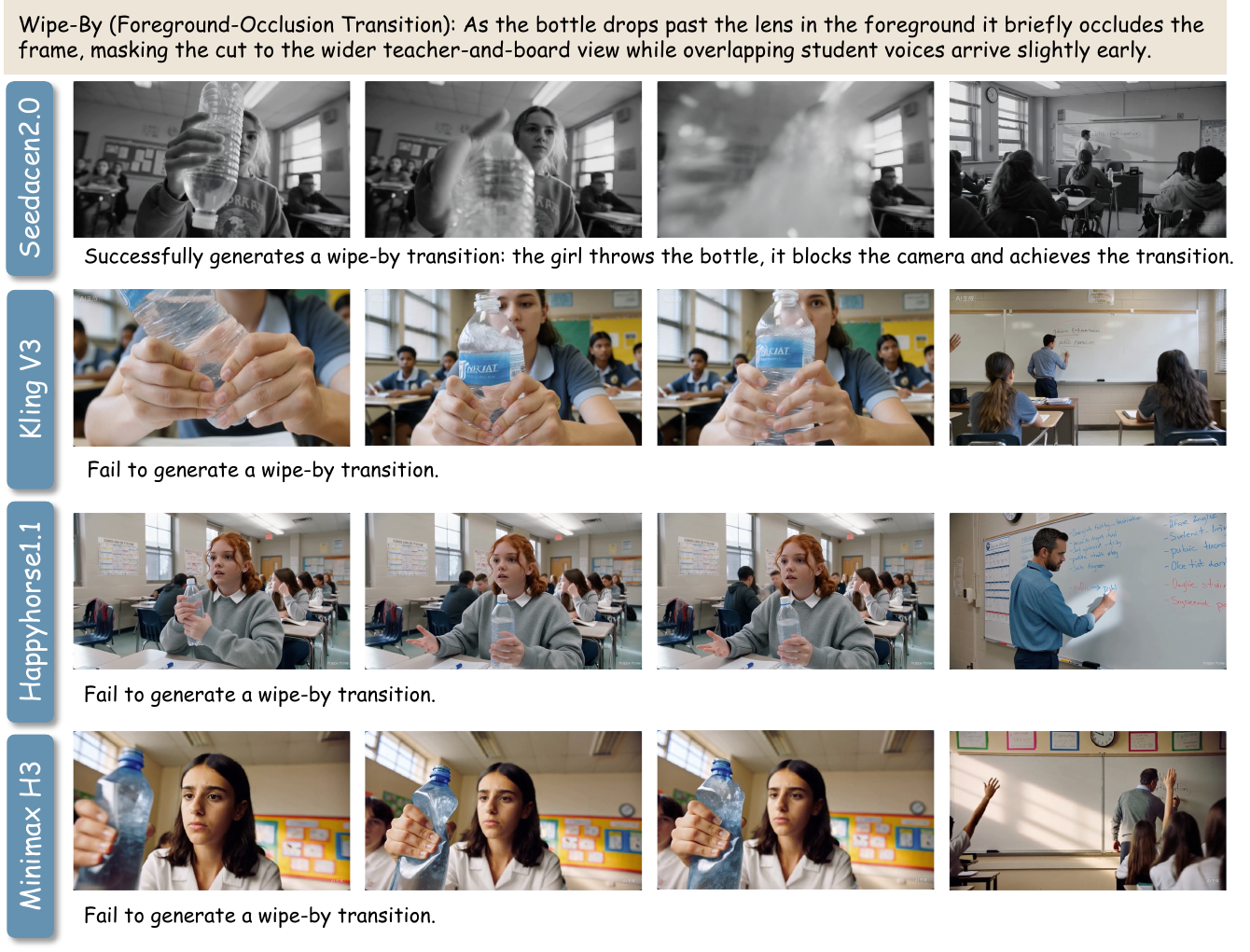}
    \caption{\textbf{Cinematographic Transition. }The prompt requires using the water bottle in the girl’s hand to create a wipe-by transition, but only Seedance2.0 is able to generate it accurately; the other models fail.
}
    \label{app:fig_morecase_transi_cine}
\end{figure}

Here we present a set of examples in which the prompt requires a wipe-by transition using the water bottle in the girl’s hand. However, most models only execute the within-shot instructions and fail to accurately carry out the transition instruction.
The results are presented in Figure \ref{app:fig_morecase_transi_cine}.

\begin{figure}[t!]
    \centering
    \includegraphics[width=0.8\linewidth]{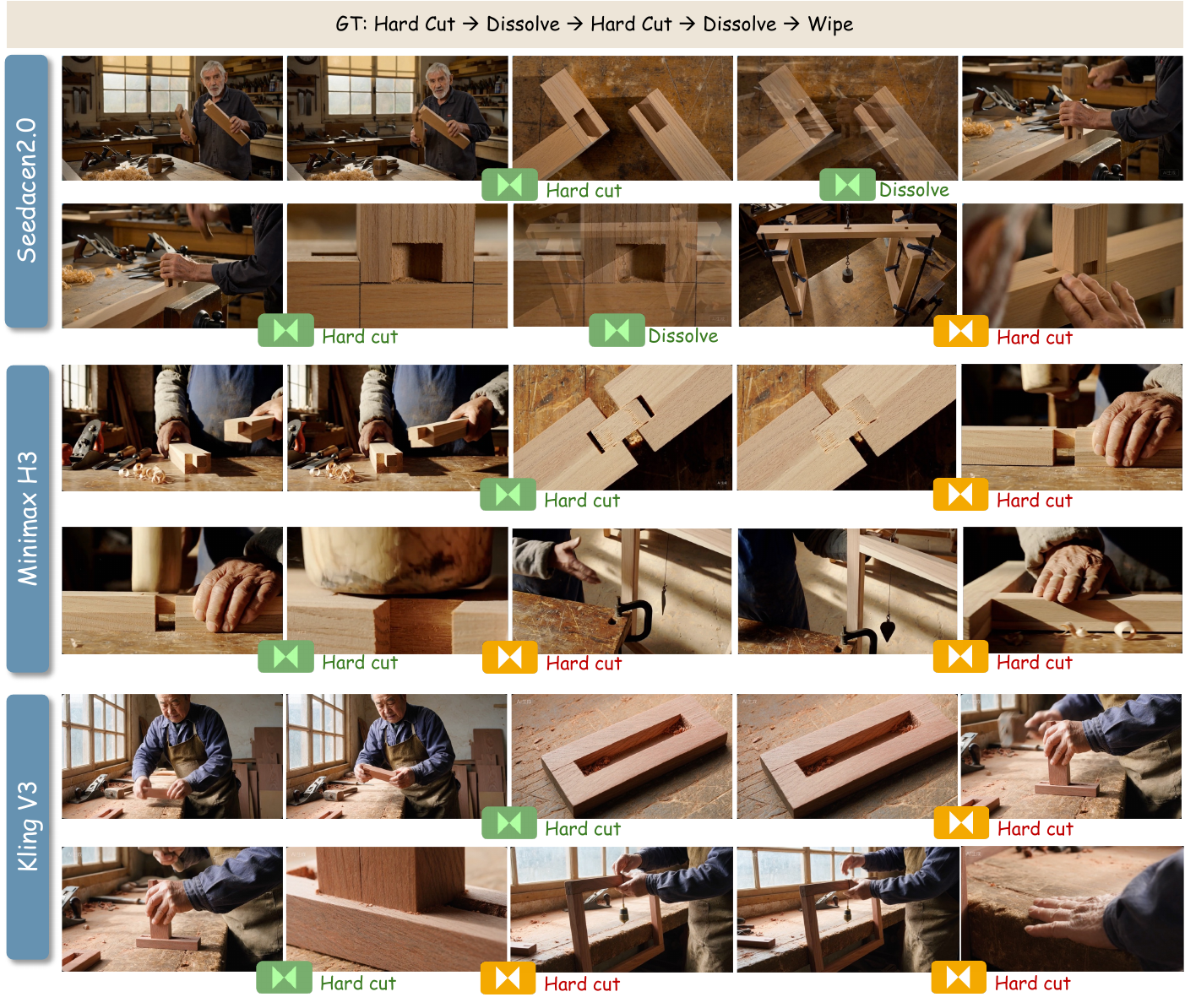}
    \caption{\textbf{Transition Optical Effect.} In complex multi-shot instructions, most models tend to collapse transition effects into hard cuts. In this example, Seedance2.0 delivers the best result.
}
    \label{app:fig_morecase_transi_opt}
\end{figure}

\paragraph{D2 Optical Transition Failure.}
Here we present a relatively complex example of a multi-shot prompt. This sample contains six shots and five transitions, covering three different types of transition effects, which makes it quite challenging for the models. From the results, it can be seen that failure cases often collapse into hard cuts and are unable to generate other types of transitions. The results are shown in Figure \ref{app:fig_morecase_transi_opt}.

\begin{figure}
    \centering
    \includegraphics[width=0.85\linewidth]{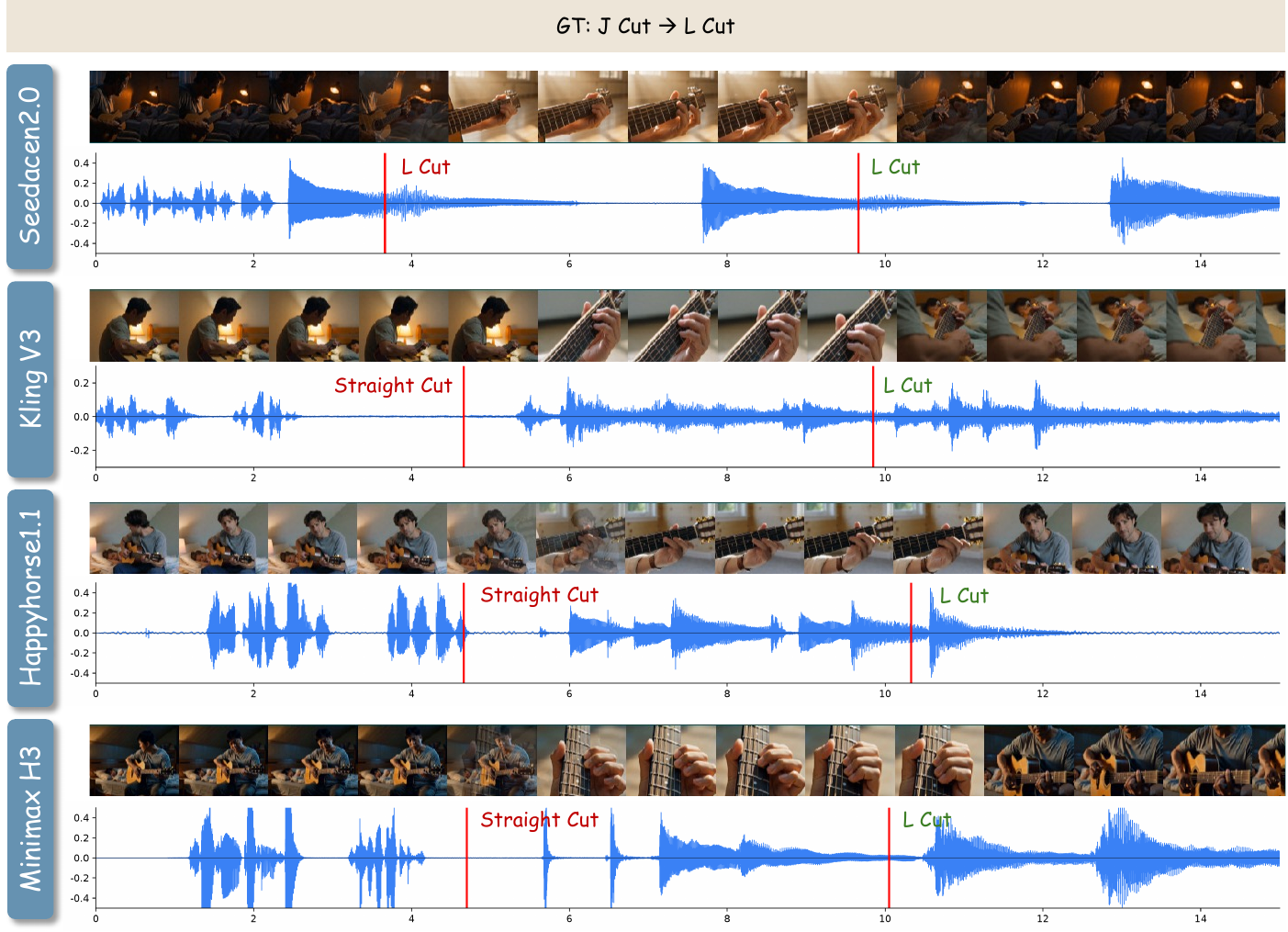}
    \caption{\textbf{Transition Audio-Video Relation. }Executing audio-video relationships across transitions is highly challenging. In this example, all models fail at the first transition.
}
    \label{app:fig_morecase_transi_av}
\end{figure}

\paragraph{D3 Audio-video Transition Relation Failure.}
We present a set of typical failure cases in audio-video relationships. The prompt requires the sound from the second shot to continue into both the first and third shots, but nearly all models fail completely at the first transition. The results are shown in Figure \ref{app:fig_morecase_transi_av}.

\begin{figure}
    \centering
    \includegraphics[width=0.85\linewidth]{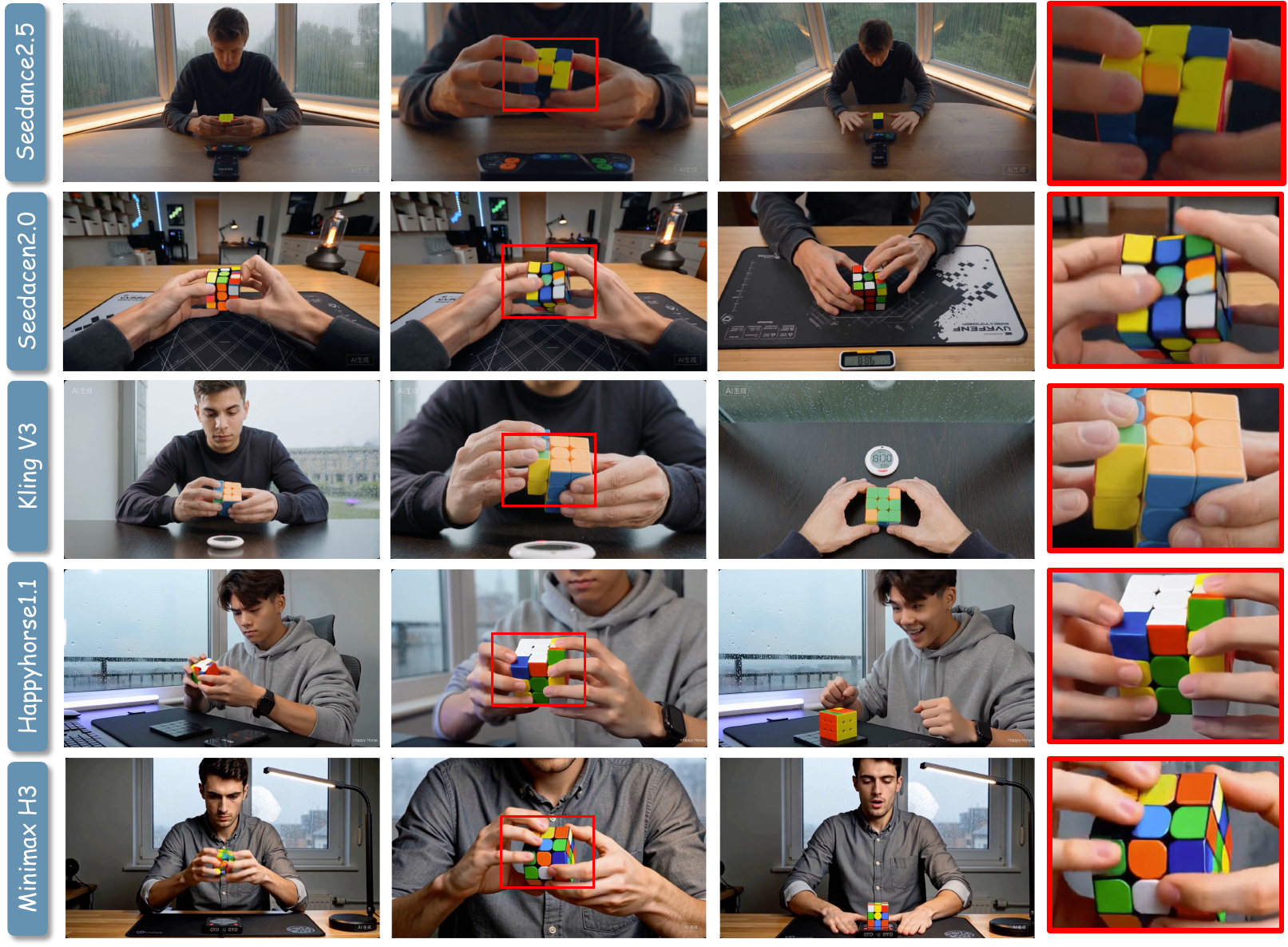}
    \caption{\textbf{Physical Consistency. }In scenes involving the generation of a Rubik’s Cube, which has a highly complex physical structure, nearly all models exhibit errors such as artifacts, distortions, and abrupt shape changes. Only Kling V3 preserves the cube’s structure relatively well, although its colors still fail to conform to common sense.
}
    \label{app:fig_morecase_phys}
\end{figure}

\paragraph{E3 Physical Consistency Failure.}
Here we present a set of relatively challenging examples involving Rubik’s Cube solving. Nearly all models fail to accurately generate scenes with such highly complex physical relationships, often producing errors such as artifacts and distortions. The results are shown in Figure \ref{app:fig_morecase_phys}.

\subsection{Training Details of OPD}
\label{app:OPD}

To obtain a compact evaluator, we apply on-policy distillation (OPD) \citep{li2026rethinking} to Qwen3-VL-8B. The pipeline contains two stages: training task-specific experts, followed by distillation into a single student model.

\paragraph{Tasks and data.}
We use four evaluation tasks derived from the human annotation protocol: free-form montage classification (\texttt{C1\_pathway\_II}), multiple-choice montage classification (\texttt{C1\_pathway\_III}), transition cinematography (\texttt{D1\_transition\_cine}), and audio-video relation classification (\texttt{D3\_audio\_video\_relation}). 
For \texttt{C1\_pathway\_III} and \texttt{D3}, we additionally include the same evidence from the expert models, including cross-shot visual similarity and audio-related signals. 

\paragraph{Expert training.}
We train one expert per task, initialized from Qwen3-VL-8B-Instruct with a LoRA adapter. The backbone is loaded in 4-bit NF4, and only LoRA parameters are optimized. We use LoRA rank 32, scaling 64, dropout 0.05, and target the standard projection layers in the transformer blocks.

Experts are trained with a GRPO-style objective. For each prompt, the model samples multiple completions, and rewards are assigned based on answer correctness, canonical label match, output validity, and formatting quality. For the multiple-choice montage task, we additionally use soft reward from annotator vote distributions. 
To mitigate imbalance, we apply class-balanced sampling on long-tail tasks and keep the best checkpoint according to validation accuracy. The main settings are: maximum prompt length 8192, maximum completion length 64, gradient accumulation 8, and AdamW optimizer; across rounds, the learning rate is varied in $[3\times10^{-6}, 10^{-5}]$, with 8--12 sampled generations per prompt.

\paragraph{On-policy distillation.}
We then distill the experts into a fresh Qwen3-VL-8B student with a larger LoRA adapter (rank 128, scaling 256, dropout 0.05), trained in bf16 with gradient checkpointing. Training data are mixed uniformly across the four tasks. At each step, with probability $1-\alpha$, the student generates an on-policy completion and is trained by reverse KL to the corresponding expert on the generated tokens. With probability $\alpha$, we instead attach the gold answer and optimize a mixture of cross-entropy and KL. Formally,
\[
\mathcal{L}=
\begin{cases}
\lambda_{\mathrm{KL}}\,\mathrm{KL}(p_s \,\|\, p_t), & \text{on-policy step},\\
\lambda_{\mathrm{CE}}\,\mathcal{L}_{\mathrm{CE}} + \lambda_{\mathrm{A}}\,\mathrm{KL}(p_s \,\|\, p_t), & \text{anchor step}.
\end{cases}
\]
We set the anchor probability $\alpha=0.25$, learning rate to $10^{-5}$, maximum prompt length to 8192, maximum completion length to 48, and gradient accumulation to 8.

\section{Ethics and Licensing}

Our CutCraft text prompts were constrained during LLM-based generation and further manually filtered.
They do not contain sensitive information, political content, violence, pornography, or other prohibited material, and do not use real personal names or other identifying information.
In addition, the videos generated in CutCraft are used solely for benchmarking purposes and for no other use.

\section{Limitations}
CutCraft is designed as a high-fidelity benchmark for editing-technique execution, but this design also brings several limitations. The evaluation pipeline is relatively long, requiring over 120s per sample on average due to shot alignment, expert-model extraction, and editing-aware multimodal judgment. Such a multi-stage design may introduce cumulative error. To reduce this risk, we have validated the major stages of the pipeline against human annotations, including shot alignment and the main editing-related dimensions, and found strong agreement with expert judgments, suggesting that the resulting evaluation noise is limited in practice. 

In addition, some metrics rely on VLM-based judgment. While this dependence is difficult to avoid for constructs such as montage, causal editing logic, and audio-video transition relations, it also makes evaluation sensitive to the quality of the judge model. We therefore performed explicit human-correlation studies and further trained an OPD-based evaluator to approximate the behavior of the stronger closed-source judge. Finally, our agentic editing system is only a baseline meant to illustrate one promising optimization direction. It is not intended as a SOTA generation solution, and improving such editing-aware generation strategies remains an important direction for future work.

\end{document}